%% file: main.tex
\documentclass[10pt,journal,compsoc]{IEEEtran}

\input{packages}
\input{authors}

\input{definitions}

\begin{document}

\title{Pre-Trained Models for Heterogeneous Information Networks}

\IEEEtitleabstractindextext{%
\begin{abstract}
	In \acl{NRL} we learn how to represent \aclp{HIN} in a low-dimen\-sional space so as to facilitate effective search, classification, and prediction solutions. Previous \acl{NRL} methods typically require sufficient task-specific labeled data to address domain-specific problems. The trained model usually cannot be transferred to out-of-domain datasets. We propose a self-supervised pre-training and fine-tuning framework, \mtv, to capture the features of a \acl{HIN}. Unlike traditional \acl{NRL} models that have to train the entire model all over again for every downstream task and dataset, \mtv only needs to fine-tune the model and a small number of extra task-specific parameters, thus improving model efficiency and effectiveness. During pre-training, we first transform the neighborhood of a given node into a sequence.
	\mtv is pre-trained based on two self-supervised tasks, masked node modeling and adjacent node prediction.
	We adopt deep bi-directional transformer encoders to train the model, and leverage factorized embedding parameterization and cross-layer parameter sharing to reduce the parameters.
	In the fine-tuning stage, we choose four benchmark downstream tasks, i.e., link prediction, similarity search, node classification, and node clustering.
	\mtv consistently and significantly outperforms state-of-the-art alternatives on each of these tasks, on four datasets.
\end{abstract}

}

\maketitle

\acresetall

\input{sections/01-intro}
\input{sections/02-related}
\input{sections/03-model}
\input{sections/04-exp}
\input{sections/05-results-and-analysis}
\input{sections/06-conclusion}

\section*{Acknowledgment}
This work was partially supported by 
NSFC under grants No. 61872446, 
NSF of Hunan Province under grant No. 2019JJ20024, and
The Science and Technology Innovation Program of Hunan Province under grant No. 2020RC4046, and
by the Hybrid Intelligence Center, 
a 10-year program
funded by the Dutch Ministry of Education, Culture and Science through 
the Netherlands
Organisation for Scientific Research, 
\url{https://hybrid-intelligence-centre.nl}.
All content represents the opinion of the authors, which is not necessarily shared or endorsed by their respective employers and/or sponsors.

\bibliographystyle{IEEEtranN}
\bibliography{references}

\begin{IEEEbiography}[{\includegraphics[width=1in,height=1.25in,clip,keepaspectratio]{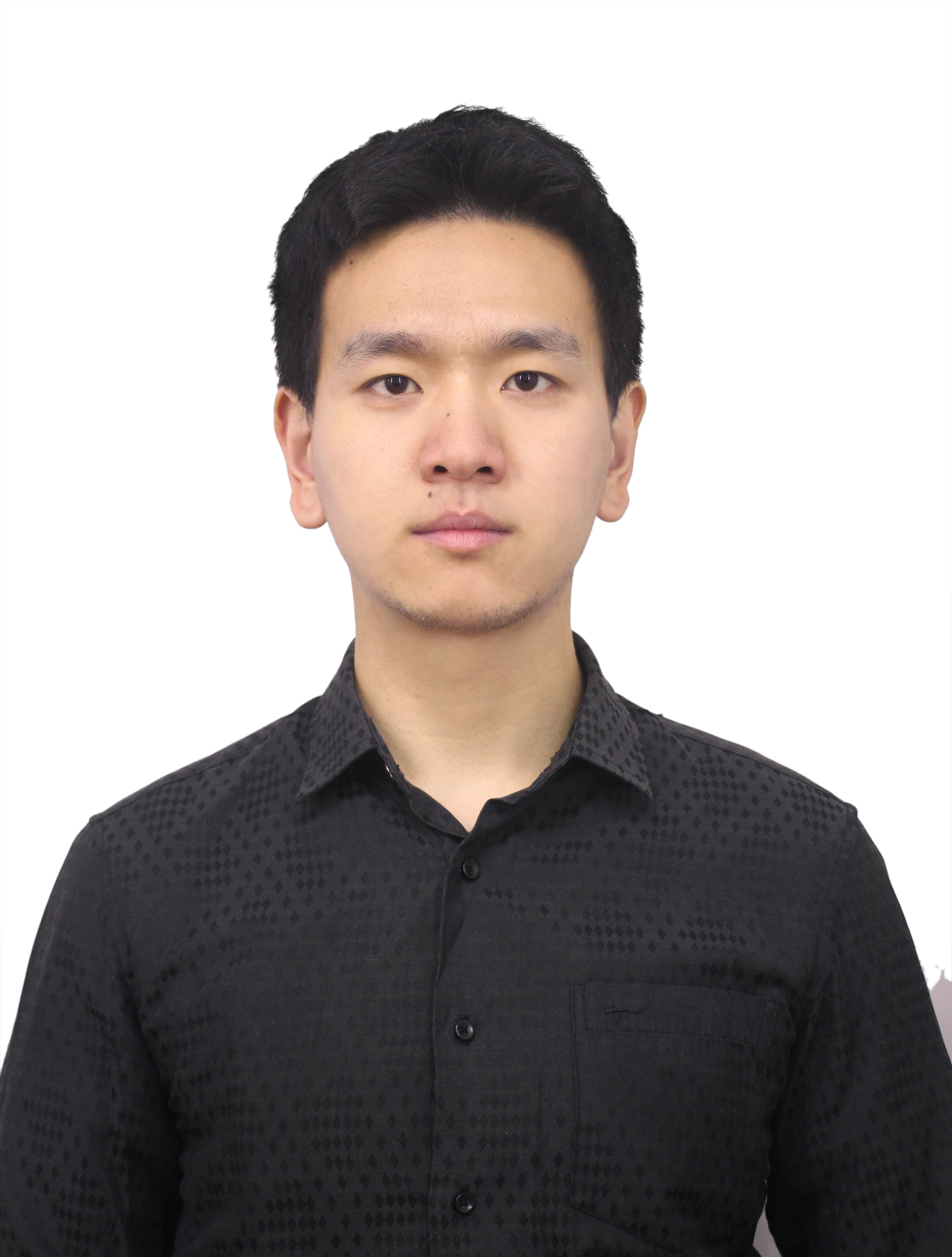}}]{Yang Fang}
	 received the Master degree from National University of Defense Technology (NUDT), China, in 2018. He is currently working toward the Ph.D. degree at NUDT, and his research mainly focuses on graph/network representation learning. 
\end{IEEEbiography}

\begin{IEEEbiography}[{\includegraphics[width=1in,height=1.25in,clip,keepaspectratio]{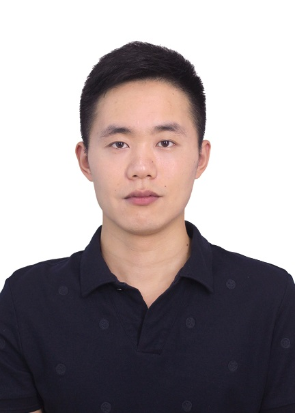}}]{Xiang Zhao}
	received the PhD degree from The University of New South Wales, Australia, in 2014. He is currently an associate professor at National University of Defense Technology, China, and he is the head of the knowledge systems engineering group. His research interests include graph data management and mining, with a special focus on knowledge graphs. 
\end{IEEEbiography}

\begin{IEEEbiography}[{\includegraphics[width=1in,height=1.25in,clip,keepaspectratio]{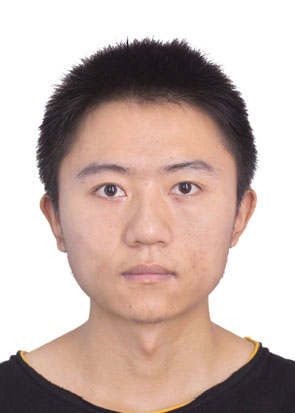}}]{Yifan Chen}
	 received the PhD degree from the University of Amsterdam (UvA), the Netherlands, in 2019, and from the National University of Defense Technology (NUDT), China, in 2020. His research interests include graph data mining and recommender systems. 
\end{IEEEbiography}

\begin{IEEEbiography}[{\includegraphics[width=1in,height=1.25in,clip,keepaspectratio]{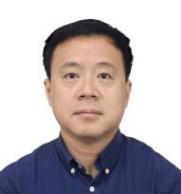}}]{WEIDONG XIAO} was born in Harbin, China,in 1968. He received the Ph.D. degree from the National University of Defense Technology(NUDT), China. He is currently a Full Professor in Science and Technology on Information Systems Engineering Laboratory, National University of Defense, China. His research interests include big data analytics and social computing.
\end{IEEEbiography}

\begin{IEEEbiography}[{\includegraphics[width=1in,height=1.25in,clip,keepaspectratio]{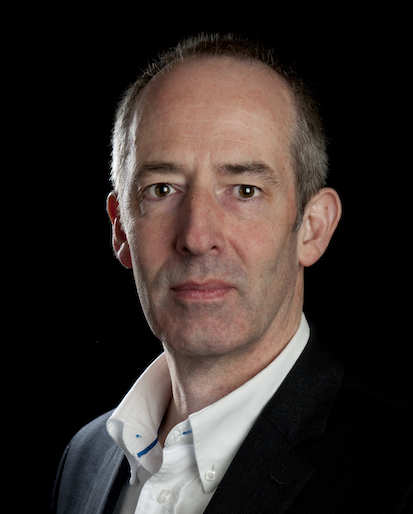}}]{Maarten de Rijke}
is a Distinguished University Professor in Artificial Intelligence and Information Retrieval at the University of Amsterdam. He is the scientific director of the national Innovation Center for Artificial Intelligence and an elected member of the Royal Netherlands Academy of Arts and Sciences. Together with a team of PhD students and postdocs he carries out research on intelligent technology to connect people to information.
\end{IEEEbiography}
\vfill

\end{document}

%% file: packages.tex
\usepackage[sort&compress,numbers]{natbib}
\usepackage{amsmath,amssymb,amsfonts}
\usepackage{algorithmic}
\usepackage{graphicx}
\usepackage{textcomp}
\usepackage{xcolor}
\usepackage{multirow} 
\usepackage{xcolor}
\usepackage[ruled,linesnumbered,boxed]{algorithm2e}
\usepackage{dsfont}
\usepackage{booktabs} 
\usepackage{xspace}
\usepackage{subfigure}
\usepackage[skip=0pt]{caption}

\usepackage{acronym}
\usepackage{url}
\usepackage{latexsym}
\usepackage[inline]{enumitem}

%% file: authors.tex
\author{Yang~Fang, Xiang~Zhao, Yifan~Chen, Weidong~Xiao,  Maarten~de~Rijke 
	\thanks{Yang Fang, Xiang Zhao, Weidong Xiao, Yifan Chen are with the National University of Defense Technology, China.		
		E-mail:  fangyang12, xiangzhao, yfchen,wdxiao@nudt.edu.cn.}
	\thanks{Maarten de Rijke is with the University of Amsterdam, the Netherlands. Email: m.derijke@uva.nl.}
	\thanks{Manuscript received XXX XX, XXXX; revised XXX XX, XXXX.}}

%% file: definitions.tex
\def\BibTeX{{\rm B\kern-.05em{\sc i\kern-.025em b}\kern-.08em
    T\kern-.1667em\lower.7ex\hbox{E}\kern-.125emX}}
\def\thepage{\arabic{page}}
\newcommand{\mtv}{\textsf{PF-HIN}\xspace}

\acrodef{ANP}{adjacent node prediction}
\acrodef{CNN}{convolutional neural network}
\acrodef{GCN}{graph convolutional network}
\acrodef{GNN}{graph neural network}
\acrodef{GCC}{graph contrastive coding}
\acrodef{HIN}{heterogeneous information network}
\acrodef{MNM}{masked node modeling}
\acrodef{NRL}{network representation learning}
\acrodef{OAG}{open academic graph}



%% file: sections/01-intro.tex

\section{Introduction}
Complex information often involves multiple types of objects and relations.
Such information can be represented via heterogeneous information networks (\acsp{HIN}\acused{HIN})~\cite{reinanda-2020-knowledge}. In a \ac{HIN} different types of nodes (objects) are connected by edges (relations)~\cite{DBLP:series/synthesis/2012Sun}.
Compared to homogeneous networks that only feature a single type of node, \acp{HIN} provide a richer modeling tool, leading to more effective solutions for search, classification, and prediction tasks~\cite{DBLP:journals/tkde/CaiZC18}.

In order to mine the rich information captured by a \ac{HIN}, \acfi{NRL} embeds a network into a low-dimen\-sional space. \ac{NRL} has drawn a significant amount of interest from the research community. Classical network embedding models like DeepWalk~\cite{DBLP:conf/kdd/PerozziAS14}, LINE~\cite{DBLP:conf/www/TangQWZYM15} and node2vec~\cite{DBLP:conf/kdd/GroverL16} have been devised for homogeneous networks, using random walks to capture the structure of networks. However, these methods lack the ability to capture a \emph{heterogeneous} information network with multiple types of objects and relations. Hence, models designed specifically for \acp{HIN} have been proposed~\cite{DBLP:conf/kdd/DongCS17,DBLP:journals/corr/HuangM17,DBLP:conf/cikm/FuLL17}. A central concept here is that of a \emph{metapath}, which is a sequence of node types with edge types in between. To leverage the relationship between nodes and metapaths, different mechanisms have been proposed, such as the heterogeneous SkipGram~\cite{DBLP:conf/kdd/DongCS17}, proximity distance~\cite{DBLP:journals/corr/HuangM17}, and the Hardmard function~\cite{DBLP:conf/cikm/FuLL17}.
Because of the limited ability of metapaths to capture the neighborhood structure of a node, the performance of these \ac{NRL} methods is limited.

Recently, \acp{GNN} have shown promising results on modeling the structure of a network~\cite{DBLP:conf/iclr/KipfW17,DBLP:conf/iclr/VelickovicCCRLB18,DBLP:conf/kdd/ZhangSHSC19}. \acp{GNN} usually involve  encoders that are able to explore and capture the neighborhood structure around a node, 
thus improving the performance on representing an \ac{HIN}. However, \acp{GNN} need to be trained in an end-to-end manner with supervised information for a task, and the model learned on one dataset cannot easily be transferred to other, out-of-domain datasets. For different datasets and tasks, the methods listed above need to be re-trained all over. Additionally, especially on large-scale datasets, the amount of available labeled data is rarely sufficient for effective training.

Inspired by advances in pre-training frameworks in language technology~\cite{DBLP:conf/nips/DaiL15,DBLP:conf/naacl/PetersNIGCLZ18,DBLP:conf/acl/RuderH18}, there is a trend to investigate pre-trained models for \ac{NRL}. In particular, \ac{GCC}~\cite{DBLP:conf/kdd/QiuCDZYDWT20} and GPT-\ac{GNN}~\cite{DBLP:conf/kdd/HuDWCS20} are the most advanced solutions in this stream.\footnote{The experiments on downstream tasks using GPT-\ac{GNN}~\cite{DBLP:conf/kdd/HuDWCS20} were conducted on the same dataset that had been employed for pre-training.} Nevertheless, they are mainly proposed for generic \ac{NRL}, meaning that they overlook the heterogeneous features of \acp{HIN}; while they are generally applicable to \acp{HIN}, they tend to fall short when handling \acp{HIN} (as demonstrated empirically in Section~\ref{sec:results-and-analysis} below).

We aim to overcome the shortcomings listed above, and propose 
\begin{enumerate*}
\item to pre-train a model on large datasets using self-supervision tasks, and
\item for a specific downstream task on a specific dataset, to use fine-tuning techniques with few task-specific parameters, so as to improve the model efficiency and effectiveness.
\end{enumerate*}
We refer to this two-stage (\textbf{P}re-training and \textbf{F}ine-tuning) framework for exploring the features of a \textbf{HIN} as \mtv. 

Given a node in a \ac{HIN}, we first explore the node's neighborhood by transforming it into a sequence to better capture the features of neighboring structure. Then, a ranking of all the nodes 
is established based on their betweenness centrality, eigencentrality and closeness centrality. We use rank-guided heterogeneous walks to generate the sequence and group different types of nodes into so-called mini-sequences, that is, sequences of nodes of the same type~\cite{DBLP:conf/kdd/ZhangSHSC19}.

We design two tasks for pre-training \mtv. One is the \acfi{MNM} task, in which a certain percentage of nodes in the mini-sequences are masked and we need to predict those masked nodes. This operation is meant to help \mtv learn type-specific node features. The other task is the \acfi{ANP} task, which is meant to capture the relationship between nodes. Given a node $u_i$ having sequence $X_i$, our aim is to predict whether the node $u_j$ with sequence $X_j$ is an adjacent node. We adopt two strategies to reduce the parameters to further improve the efficiency of \mtv, i.e., factorized embedding parameterization and cross-layer parameter sharing. The large-scale dataset we use for pre-training is the \ac{OAG}, containing 179 million nodes and 2 billion edges.

During fine-tuning, we choose four benchmark downstream tasks:
\begin{enumerate*}
\item link prediction,
\item similarity search,
\item node classification, and
\item node clustering.
\end{enumerate*}
In link prediction and similarity search, we use node sequence pairs as input, and identify whether there is a link between two nodes and measure the similarity between two nodes, respectively. In the node classification and node clustering tasks, we use a single node sequence as input, employing a softmax layer for classification and a k-means algorithm for clustering, respectively.

In our experiments, which are meant to demonstrate that \mtv is transferable across  datasets, besides a subset of \ac{OAG} denoted as \ac{OAG}-mini, we include three other datasets for downstream tasks: DBLP, YELP and YAGO. \mtv consistently and significantly outperforms the state-of-the-art on these downstream tasks.

Our main contributions can be summarized as follows:

\begin{itemize}[nosep,leftmargin=*]
	\item We propose a pre-training and fine-tuning framework \mtv to mine information contained in a \ac{HIN}; \mtv is transferable to different downstream tasks and to datasets of different domains.
	\item We adopt deep bi-directional transformer encoders to capture the structural features of a \ac{HIN}; the architecture of \mtv is a variant of a \ac{GNN}.
	\item We use type-based masked node modeling and adjacent node prediction tasks to pre-train \mtv; both help \mtv to capture heterogeneous node features and relationships between nodes.
	\item We show that \mtv significantly outperforms the state of the art on four benchmark downstream tasks across datasets.
\end{itemize}

%% file: sections/02-related.tex

\section{Related work}
\label{sec:related}
\subsection{Network representation learning}
Research on \ac{NRL} traces back to dimensionality reduction techniques~\cite{DBLP:conf/nips/BelkinN01,Cox2008Multidimensional,Roweis2000Nonlinear,DBLP:conf/kdd/WangC016}, which utilize feature vectors of nodes to construct an affinity graph and then calculate eigenvectors. Graph factorization models~\cite{DBLP:conf/www/AhmedSNJS13} represent a graph as an adjacency matrix, and generate a low-dimensional representation via matrix factorization. Such models suffer from high computational costs and data sparsity, and cannot capture the global network structure~\cite{DBLP:conf/www/TangQWZYM15}.

Random walks or paths in a network are being used to help preserve the local and global structure of a network. DeepWalk~\cite{DBLP:conf/kdd/PerozziAS14} leverages random walks and applies the SkipGram word2vec model to learn network embeddings. node2vec~\cite{DBLP:conf/kdd/GroverL16} extends DeepWalk; it adopts a biased random walk strategy to explore the network structure. LINE~\cite{DBLP:conf/www/TangQWZYM15} harnesses first- and second-order proximities to encode local and neighborhood structure information.

The aforementioned approaches are designed for homogeneous networks; other methods have been introduced for heterogeneous networks.
PTE~\cite{DBLP:conf/kdd/TangQM15} defines the conditional probability of nodes of one type generated by nodes of another, and forces the conditional distribution to be close to its empirical distribution.
metapath2vec~\cite{DBLP:conf/kdd/DongCS17} has a heterogeneous SkipGram with its context window restricted to one specific type. HINE~\cite{DBLP:journals/corr/HuangM17} uses a metapath based proximity and minimizes the distance between nodes' joint probability and empirical probabilities. HIN2Vec~\cite{DBLP:conf/cikm/FuLL17} uses Hadamard multiplication of nodes and metapaths to capture features. 

What we contribute to \ac{NRL} on top of the work listed above is an efficient and effective method for representation learning for \acp{HIN} based on \acfp{GNN}.

\subsection{Graph neural networks}

\Ac{GNN} models have shown promising results for representing networks.
Efforts have been devoted to generalizing convolutional operations from visual data to graph data. 
\citet{DBLP:journals/corr/BrunaZSL13} propose a spectral graph theory based graph convolution operation. 
\Acp{GCN}~\cite{DBLP:conf/iclr/KipfW17} adopt localized first-order approximations of spectral graph convolutions to improve scalability. There is a line of research to improve spectral \ac{GNN} models~\cite{DBLP:conf/nips/DefferrardBV16,DBLP:journals/corr/HenaffBL15,DBLP:conf/aaai/LiWZH18,DBLP:journals/tsp/LevieMBB19}, but it processes the whole graph simultaneously, leading to efficiency bottlenecks. To address the problem, spatial \ac{GNN} models have been proposed~\cite{DBLP:conf/cvpr/MontiBMRSB17,DBLP:conf/icml/NiepertAK16,DBLP:conf/kdd/GaoWJ18,DBLP:conf/nips/HamiltonYL17}. GraphSAGE~\cite{DBLP:conf/nips/HamiltonYL17} leverages a sampling strategy to iteratively sample neighboring nodes instead of the whole graph. 
\citet{DBLP:conf/kdd/GaoWJ18} utilize a sub-graph training method to reduce memory and computational cost.

\Acp{GNN} fuse neighboring nodes or walks in graphs so as to learn a new node representation~\cite{DBLP:conf/iclr/VelickovicCCRLB18,DBLP:conf/uai/ZhangSXMKY18,DBLP:conf/kdd/LeeRK18}. The main difference with convolution based models is that graph attention networks introduce attention mechanisms to assign higher weights to more important nodes or walks. GAT~\cite{DBLP:conf/iclr/VelickovicCCRLB18} harnesses masked self-attention layers to apply different weights to different nodes in a neighborhood, to improve efficiency on graph-structured data. GIN~\cite{DBLP:conf/iclr/XuHLJ19} models injective multiset functions for neighbor aggregation by parameterizing universal multiset functions with neural networks.

The above \ac{GNN} models have been devised for homogeneous networks as they aggregate neighboring nodes or walks regardless of their types. Targeting \acp{HIN}, HetGNN~\cite{DBLP:conf/kdd/ZhangSHSC19} first samples a fixed number of neighboring nodes of a given node and then groups these based on their  types. Then, it uses a neural network architecture with two modules to aggregate the feature information of the neighboring nodes. One module is used to encode features of each type of node, the other to aggregate features of different types. HGT~\cite{DBLP:conf/www/HuDWS20} uses node- and edge-type dependent parameters to describe heterogeneous attention over each edge; it also uses a heterogeneous mini-batch graph sampling algorithm for training.

What we contribute to \acp{GNN} is that the traditional \ac{NRL} and \ac{GNN} models  listed above need to be re-trained all over again for different datasets and tasks, while our proposal \mtv only needs fine-tuning using a small number of task-specific parameters for a specific task and dataset, after pre-training via self-supervision tasks.

\subsection{Graph pre-training}

There exist relatively few approaches for pre-training a \ac{GNN} model for downstream tasks. InfoGraph~\cite{DBLP:conf/iclr/SunHV020} maximizes the mutual information between graph-level embeddings and sub-structure embeddings. \citet{DBLP:conf/iclr/HuLGZLPL20} pre-train a \ac{GNN} at the level of nodes and graphs to learn local and global features, showing performance improvements on various graph classification tasks. Our proposed model, \mtv, differs as we focus on node-level transfer learning and pre-train our model on a single (large-scale) graph.

\citet{DBLP:journals/corr/abs-1905-13728} design three pre-training tasks: denoising link reconstruction, centrality score ranking, and cluster preserving. GPT-GNN~\cite{DBLP:conf/kdd/HuDWCS20} adopts HGT~\cite{DBLP:conf/www/HuDWS20} as its base \ac{GNN} and uses attribute generation and edge generation as pre-training tasks. \citet{DBLP:conf/kdd/HuDWCS20} only conduct their downstream tasks on the same dataset that was used for pre-training. GCC~\cite{DBLP:conf/kdd/QiuCDZYDWT20} designs subgraph instance discrimination as a pre-training task and uses contrastive learning to train \acp{GNN}, with its base \acp{GNN} as GIN; then it transfers its pre-trained model to different datasets. However, it is designed for homogeneous networks, not apt to exploit heterogeneous networks. \citet{DBLP:conf/nips/HwangPKKHK20} proposes to adopt meta-path prediction as a pre-training task, which is to predict if two nodes are connected via a meta-path. The authors have not assigned a name for this method, so use MPP to refer to it in this paper.

What we contribute to graph pre-training on top of the work listed above is that we are able to not only fine-tune the proposed model across different tasks and different datasets, but can also deal with heterogeneous networks.

%% file: sections/03-model.tex
\begin{figure*}
	\centering
	\includegraphics[clip,trim=1mm 0mm 1mm 0mm, width=0.95\textwidth]{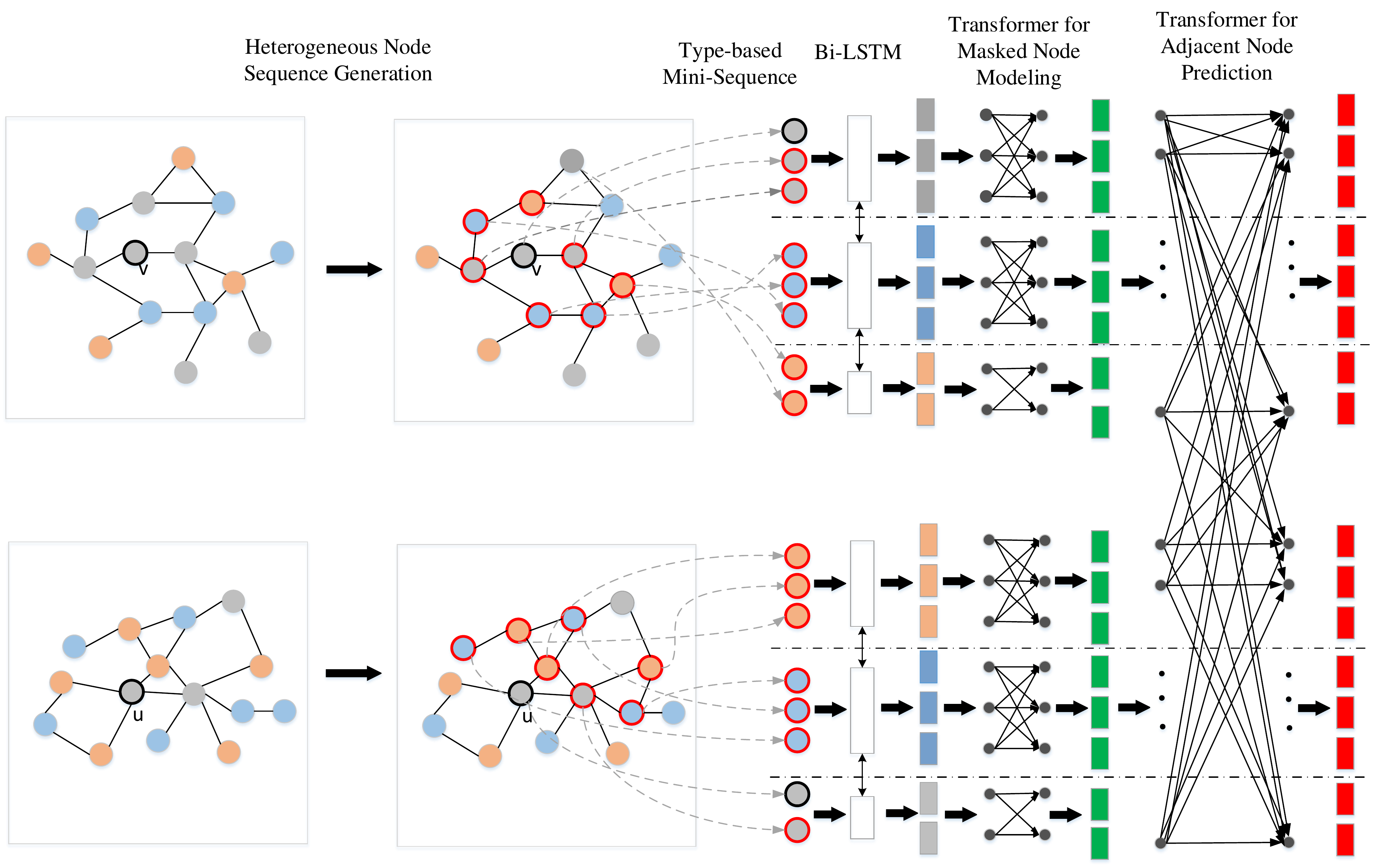}
	\caption{The pre-training procedure that makes up \mtv.  The generation of node sequences is detailed in Section~\ref{subsection:3b}. Bi-LSTM based learning of input embeddings is described in Section~\ref{subsection:3c}. \Acf{MNM} is discussed on Section~\ref{subsection:3d} and followed by \acf{ANP}.
	}
	\label{fig:frame}
\end{figure*}

\section{The Proposed Model \mtv}
\label{sec:model}

\subsection{Preliminaries}
A \acfi{HIN} is a graph $G=(V,E,T)$, where $V$ denotes the set of nodes and $E$ denotes the set of edges between nodes.
Each node and edge is associated with a type mapping function, $\phi: V \rightarrow T_V $ and $\varphi: E \rightarrow T_E$, respectively. $T_V$ and $T_E$ denote the sets of node and edge types. A \ac{HIN} is a network where $|T_V|>1$ and/or $|T_E|>1$.

A visual presentation of the proposed model, \emph{pre-training and fin-tuning \acl{HIN}} (\mtv), is given in Figure~\ref{fig:frame}.
Below, we describe the node sequence generation procedure, the input representation, followed by the pre-training and fine-tuning stages of \mtv.

\subsection{Heterogeneous node sequence generation}
\label{subsection:3b}
We first transform the structure of a node's neighborhood to a sequence of length $k$. To measure the importance of nodes based on the structural roles in the graph, node centrality is proposed in~\cite{Bonacich1987Power}. This framework makes use of three centrality metrics\footnote{This framework is extensible in the sense that additional metric that is of particular interest to the user can be explicitly supplemented after the three metrics, as long as the combination exhibits better performance in downstream tasks. In our implementation, we stick to the basic version consisted of three metrics to demonstrate the effectiveness of the framework.}, i.e.,
\begin{enumerate*}
\item betweenness centrality,
\item eigencentrality, and
\item closeness centrality.
\end{enumerate*}
Betweenness centrality is calculated as the fraction of shortest paths that pass through a given node.
Eigencentrality measures the influence of a node on its neighbors.
Closeness centrality computes the total length of the shortest paths between the given node and others.
We assign learnable  weights to these metrics.

To capture heterogeneous features of a node's neighbor, we adopt a so-called rank-guided heterogeneous walk to form a sequence. The walk is with restart; it will iteratively travel from a node to its neighbors. It starts from node $v$, and it first reaches out to a node with a higher rank, which is what makes the walk rank-guided. This walk will not stop until it collects a pre-determined number of nodes. In order to assign the model with a sense of heterogeneity, we constrain the number of different types to be collected in the sequence so that every type of node can be included. We group nodes into mini-sequences, where a mini-sequence is a sequence of nodes having the same type~\cite{DBLP:conf/kdd/ZhangSHSC19}. In each mini-sequence, the nodes are sorted based on each node's rank, which serve as a kind of position information.

Importantly, unlike traditional sampling strategies like random walks, breadth first search or depth first search, our sampling strategy is able to extract important and influential neighboring nodes for each node by selecting nodes with a higher rank; this allows us to capture more representative structural information of a neighborhood. Our sampling strategy collects all types of node for each node while traditional strategies ignore the nodes' types. Nodes of the same type are grouped in mini-sequences so that further type-based analysis can be conducted to capture the heterogeneous features of a \ac{HIN}. More empirical results with analysis are provided in Section~\ref{sec:ablation}.

\subsection{Input embeddings learned via Bi-LSTMs}
\label{subsection:3c}
\label{Initial}
After generating the sequences, we learn the input embeddings of each node in the sequence using a Bi-LSTM layer. Given the input sequence $\{x_1,x_2,\ldots,x_n\}$, in which $x_i\in\mathbb{R}^{d\times 1}$, a Bi-LSTM is used to capture the interaction relationships between nodes. The Bi-LSTM is composed of a forward and a backward LSTM layer. The LSTM layer is defined as follows:
\begin{align}
\mathbf{j}_i&=\delta(\mathbf{W}_{xj}x_i+\mathbf{W}_{hj}h_{i-1}+\mathbf{W}_{cj}c_{i-1}+\mathbf{b}_j),
\\
\mathbf{f}_{i}&=\delta(\mathbf{W}_{xf}x_i+\mathbf{W}_{hi}h_{i-1}+\mathbf{W}_{cf}c_{i-1}+\mathbf{b}_f),
\\
\mathbf{z}_{i}&=\mathrm{tanh}(\mathbf{W}_{xc}x_i+\mathbf{W}_{hc}h_{i-1}+\mathbf{b}_c),
\\
c_{i}&=\mathbf{f}_{i}\odot c_{i-1}+\mathbf{j}_i\odot\mathbf{z}_i,
\\
\mathbf{o}_i&=\delta(\mathbf{W}_{xo}x_i+\mathbf{W}_{ho}h_{i-1}+\mathbf{W}_{co}c_{i}+\mathbf{b}_o),
\\
h_{i}&=\mathbf{o}_{i}\mathrm{tanh}(c_{i}),
\end{align}
where $h_{i}\in\mathbb{R}^{d/2}\times1$ is the output hidden state of node $i$, $\odot$ represents the element-wise product, $\mathbf{W}\in \mathbb{R}^{(d/2)\times(d/2)}$ and $b\in \mathbb{R}^{d/2\times1}$ are  learnable parameters, which denote weight and bias, respectively; $\mathbf{j}_{i}$, $\mathbf{f}_{i}$, $\mathbf{o}_{i}$ are the input gate vector, forget gate vector and output vector, respectively. We concatenate the hidden state of the forward and backward LSTM layers to form the final hidden state of the Bi-LSTM layer. For each type of node, we adopt different Bi-LSTMs so as to extract type-specific features.

\subsection{Masked node modeling}
\label{subsection:3d}
After generating the input embeddings via a Bi-LSTM, we adopt \acf{MNM} as our pre-training task. We randomly mask a percentage of the input nodes and then predict those masked nodes. We conduct this task on the type-based mini-sequences generated by the aforementioned rank guided heterogeneous walk. For each group of nodes with the same type, we randomly mask nodes in the mini-sequence. Given the mini-sequence of type $t$, denoted as $\{x_1^t,x_2^t,\ldots,x_n^t\}$, we randomly choose 15\% of the nodes to be replaced. And for a chosen node $x_i^t$, we replace its token with the actual [MASK] token with 80\% probability, another random  node token with 10\% probability  and the unchanged $x_i^t$ with 10\% probability. The masked sequence is fed into the bi-directional transformer encoders. The embeddings generated via the Bi-LSTM are used as token embeddings, while the rank information is transferred as position embeddings. After the transformer module, the final hidden state $h_i^{t_L}$ corresponding to the [MASK] token is fed to a feedforward layer. The output is used to predict the target node via a softmax classification layer:
\begin{align}
z_i^t&= \mathrm{Feedforward} (h_i^{t_L}),
\label{equation 1}
\\
\mathbf{p}_i^t&= \mathrm{softmax} (\mathbf{W}^{\mathrm{MNM}}z_i^t),
\label{equation 2}
\end{align}
where $z_i^t$ is the output of the feedforward layer, $\mathbf{W}^{\mathrm{MNM}}\in V^t
\times d$ is the classification weight shared with the input node embedding matrix, $V^t$ is the number of nodes in the $t$-type mini-sequence, $d$ is the dimension of the hidden state size, $\mathbf{p}_i^t$ is the predicted distribution of $x_i^t$ over all nodes.

For training, we use the cross-entropy between the one-hot label $\mathbf{y}_i^t$ and the prediction $\mathbf{p}_{i}^t$:
\begin{equation}\label{equation1}
\mathcal{L}^t_\mathrm{MNM}=-\sum_{m}y_m^t \log p_m^t,
\end{equation}
where $y_m^t$ and $p_m^t$ are the $m$-th components of $\mathbf{y}_i^t$ and $\mathbf{p}_i^t$, respectively. We adopt a smoothing strategy by setting $y_m^t=\epsilon$ for the target node and $y_m^t=\frac{1-\epsilon}{V^t-1}$ for each of the other nodes. By doing so, we loosen the restriction that a one-hot label corresponds to only one answer.

\subsection{Adjacent node prediction}
\label{subsection:3e}
\label{ANP}
After the masked node modeling module, we design another pre-training task, i.e., \acf{ANP}, to further capture the relationship between nodes. Unlike the \ac{MNM} task, which operates on type-based mini-sequences, we perform the \ac{ANP} task on full sequences, and we compare two full sequences to see whether their starting nodes are adjacent or not. The reason that we do not perform the \ac{ANP} task on type-based mini-sequences is that given $k$ types of nodes, there will be $k(k-1)/2$ mini-sequence pairs to be analyzed, which is very time-consuming.

In our setting, for node $v$ with sequence $X_v$ and node $u$ with sequence $X_u$, 50\% of the time we choose $u$ to be the actual adjacent node of $v$ (labeled as IsAdjacent), and 50\% of the time we randomly choose $u$ from the corpus (labeled as NotAdjacent). Given the classification layer weights $\mathrm{W}^{\mathrm{ANP}}$, the scoring function $s_\tau$ of whether the node pair is adjacent is shown as follows:
\begin{equation}\label{equation 4}
s_\tau=\mathrm{sigmoid}(C\mathrm{W}^{\mathrm{ANP}^{T}}),
\end{equation}
where $s_\tau\in \mathbb{R}^2$ is a binary vector with $s_{\tau0},s_{\tau1}\in [0,1]$ and $s_{\tau0}+s_{\tau1}=1$, $C\in \mathbb{R}^H$ denotes the hidden vector of classification label used in a transformer architecture~\cite{DBLP:conf/naacl/DevlinCLT19}.
Considering the positive adjacent node pair $\mathbb{S}^{+}$ and a negative adjacent node pair $\mathbb{S}^{-}$, we calculate a cross-entropy loss as follows:
\begin{equation}\label{equation2}
\mathcal{L}_\mathrm{ANP}=-\sum_{\tau \in \mathbb{S}^{+} \cup \mathbb{S}^{-}}(y_\tau\mathrm{log}(s_{\tau0}))+(1-y_\tau)\mathrm{log}(s_{\tau1})),
\end{equation}
where $y_\tau$ is the label (positive or negative) of that node pair.

\subsection{Transformer architecture}
\label{subsection:3f}
Our two pre-training tasks share the same transformer architecture. To increase the training speed of our model, we adopt two parameter reduction techniques to lower memory requirements, inspired by the ALBERT architecture~\cite{DBLP:conf/iclr/LanCGGSS20}. Instead of setting the node embedding size $Q$ to be equal to the hidden layer size $H$ like BERT does, we make more efficient use of the total number of model parameters, dictating that $H\gg Q$. We adopt factorized embedding parameterization, which  decomposes the parameters into two smaller matrices. Rather than mapping the one-hot vectors directly to a hidden space with size $H$, we first map them to a low-dimensional embedding space with size $Q$, and then map it to the hidden space. Additionally, we adopt cross-layer parameter sharing to further boost efficiency. Traditional sharing mechanisms either only share the feed forward network parameters across layers or only the attention parameters. We share \emph{all} parameters across layers.

We denote the number of transformer layers as $L$, and the number of self-attention heads as $A$. For our parameter settings we follow the configuration of ALBERT~\cite{DBLP:conf/iclr/LanCGGSS20}, where $L$ is set to 12, $H$ to 768, $A$ to 12, $Q$ to 128, and the total number of parameters is equal to 12M.
For the procedure of our pre-training task, see Algorithm~\ref{algorithm1}.
\IncMargin{1em}
\begin{algorithm}[t]	
	\SetAlgoNoLine
	\SetKwInOut{Input}{\textbf{Input}}\SetKwInOut{Output}{\textbf{Output}} 	
	\Input{
		
		Input HIN $G$\;}
	
	\Output{
		
		Optimized model parameters $\Theta$ (for downstream tasks)\;\\
	}
	\BlankLine	
	Generates the node sequences via rank guided heterogeneous walk\;
	\For {each pair of sampled sequences}
	{Apply Bi-LSTM on type-based mini-sequences to learn the input embeddings of each node in the sequence\;
		\For {each sequence}
		{Mask nodes in the type-based mini-sequences\;
			Feed the mini-sequences into transformer layers\;
			Calculate the masked node modeling loss by Eq.(\ref{equation1})\;}
		Feed the two sequenes into transformer layers\;
		Calculate the ajacent node prediction loss by Eq.(\ref{equation2})\;
		Update the parameters $\Theta$ by Adam.
	}

	\Return Optimized pre-trained model parameters $\Theta^*$.	
	\caption{The Pre-training procedure of \mtv.}\label{algorithm1}
\end{algorithm}
\DecMargin{1em}

\subsection{Fine-tuning \mtv}
\label{subsection:3g}
The self-attention mechanism in the transformer allows \mtv to model many downstream tasks. Fine-tuning can be realized by simply swapping out the proper inputs and outputs, regardless of the single node sequence or sequence pairs used. For each downstream task, the task-specific inputs and outputs are simply plugged into \mtv and all parameters are fine-tuned end-to-end.
Here, we introduce four tasks:
\begin{enumerate*}
\item link prediction,
\item similarity search,
\item node classification, and
\item node clustering as downstream tasks.
\end{enumerate*}

Specifically, in link prediction, we predict whether there is a link between two nodes, and the inputs are the node sequence pairs. To generate output, we feed the classification label into the sigmoid layer, so as to predict the existence of a link between two nodes. The only new parameters are the classification layer weights $W\in \mathbb{R}^{2\times H}$, where $H$ is the size of hidden state.

In similarity search, in order to measure the similarity between two nodes, we use the node sequence pairs as input. We leverage the token-level output representations to compute the similarity score of two nodes.

In node classification, we only use a single node sequence as input and  generate the classification label via a softmax layer. To calculate the classification loss, we only need to add classification layer weights $W\in \mathbb{R}^{K\times H}$ as new parameters, where $K$ is the number of classification labels and $H$ is the size of hidden state.

In node clustering, we also use a one node sequence as input and then put the token-level output embeddings to a clustering model, so as to cluster the data.

Experimental details for these downstream tasks are introduced in Section~\ref{sec:exp} below.

%% file: sections/04-exp.tex

\section{Experimental Setup}
\label{sec:exp}
\label{exps}
We detail our datasets, baseline models, and parameter settings.

\subsection{Datasets}\label{dataset}
We adopt the open academic graph OAG\footnote{\url{https://www.openacademic.ai/oag/}} as our pre-training dataset, which is a heterogeneous academic dataset. It contains over 178 million nodes and 2.223 billion edges with five types of node:
\begin{enumerate*}
\item author,
\item paper,
\item venue,
\item field, and
\item institute.
\end{enumerate*}

For downstream tasks, we transfer our pre-trained model to four datasets:
\begin{enumerate*}
\item OAG-mini,
\item DBLP,
\item YELP, and
\item YAGO.
\end{enumerate*}
OAG-mini is a subset extracted from OAG; the authors are split into four areas: machine learning, data mining, database, and information retrieval.  DBLP\footnote{\url{http://dblp.uni-trier.de}} is also an academic dataset with four types of node:
\begin{enumerate*}
\item author,
\item paper,
\item venue, and
\item topic;
\end{enumerate*}
the authors are split into the same areas as those in OAG-mini. YELP\footnote{\url{https://www.yelp.com/dataset\_challenge}} is a social media dataset, with restaurants reviews and four types of node:
\begin{enumerate*}
\item review,
\item customer,
\item restaurant, and
\item food-related keywords.
\end{enumerate*}
The restaurants are separated into
\begin{enumerate*}
\item Chinese food,
\item fast food, and
\item sushi bar.
\end{enumerate*}
YAGO\footnote{\url{https://old.datahub.io/dataset/yago}} is a knowledge base and we extracted a subset of it containing movie information, having five types of node:
\begin{enumerate*}
\item movie,
\item actor,
\item director,
\item composer, and
\item producer.
\end{enumerate*}
The movies are split into five types:
\begin{enumerate*}
\item action,
\item adventure,
\item sci-fi,
\item crime, and
\item horror.
\end{enumerate*}
The dataset statistics are shown in Table~\ref{tab:statictics}.

\begin{table} [t]
	\centering
	\caption{Dataset statistics.}
	\label{tab:statictics}
	\begin{tabular}{lrrcc}
		\toprule
		Dataset & \#nodes & \#edges  &\#node types   \\
		\midrule
		OAG &178,663,927 & 2,236,196,802&5\\
		OAG-mini & 473,324&2,343,578 &5 	\\	
		DBLP   & 301,273 & 1,382,587 & 4  \\
		YELP   & 201,374 & 872,432 & 4   \\
		YAGO   & 52,384 &143,173 &5\\
		\bottomrule
	\end{tabular}
\end{table}

\subsection{Algorithms used for comparison}
We first choose network embedding methods to directly train the downstream datasets for specific tasks as baselines:  DeepWalk~\cite{DBLP:conf/kdd/PerozziAS14}, LINE~\cite{DBLP:conf/www/TangQWZYM15} and node2vec~\cite{DBLP:conf/kdd/GroverL16}; they were originally applied to  homogeneous information networks. DeepWalk and node2vec leverage random walks, while node2vec uses a biased walk strategy to  capture the network structure. LINE uses the local and neighborhood structural information via first-order and second-order proximities.

We include three state-of-the-art algorithms devised for \acp{HIN}: metapath2vec~\cite{DBLP:conf/kdd/DongCS17}, HINE~\cite{DBLP:journals/corr/HuangM17}, HIN2Vec~\cite{DBLP:conf/cikm/FuLL17}. They are all based on metapaths, but differ in the way they use metapath features: metapath2vec adopts heterogeneous SkipGrams, HINE proposes a metapath-based notion of proximity, and HIN2Vec utilizes the Hadamard multiplication of nodes and metapaths.

We also include other GNN models, i.e., GCN~\cite{DBLP:conf/iclr/KipfW17}, GAT~\cite{DBLP:conf/iclr/VelickovicCCRLB18}, GraphSAGE~\cite{DBLP:conf/nips/HamiltonYL17} and GIN~\cite{DBLP:conf/iclr/XuHLJ19}, which were originally devised for homogeneous information networks. GCN and GraphSAGE are based on convolutional operations, while GCN requires the Laplacian of the full graph, and GraphSAGE only needs a node's local neighborhood. GAT employs an attention mechanism to capture the correlation between central node and neighboring nodes. GIN uses parameterizing universal multiset functions with neural networks to  model injective multiset functions for neighbor aggregation.

We also select HetGNN~\cite{DBLP:conf/kdd/ZhangSHSC19}, HGT~\cite{DBLP:conf/www/HuDWS20} as models for comparison; both have been devised for \ac{HIN} embeddings. HetGNN samples heterogeneous neighbors, grouping them based on their node types, and then aggregates feature information of the sampled neighboring nodes. HGT has node- and edge-type dependent parameters to characterize heterogeneous attention over each edge.

Aside from those network embedding methods, for a fair comparison, we also choose GPT-GNN~\cite{DBLP:conf/kdd/HuDWCS20}, GCC~\cite{DBLP:conf/kdd/QiuCDZYDWT20} and MPP~\cite{DBLP:conf/nips/HwangPKKHK20} to run the entire pre-training and fine-tuning pipeline. GPT-GNN utilizes attribute generation and edge generation tasks to pre-train GNN, with HGT as its base GNN. GCC adopts subgraph instance discrimination as a pre-training task, taking GIN as its base GNN. MPP proposes to adopt meta-path prediction as the pre-training task.

\subsection{Parameters}
For pre-training, we set the generated sequence length $k$ to 20. The dimension of the node embedding is set to 128 and the size of hidden state is set to 768. On transformer layers, we use 0.1 as the dropout probability. The Adam learning rate is initiated as 0.001 with a linear decay. We use 256 sequences to form a batch and the training epoch is set to 20. The training loss is the sum of the mean masked node modeling likelihood and the mean adjacent node prediction likelihood.

In fine-tuning, most parameters remain the same as  in pre-training, except the learning rate, batch size and number of epochs. We use grid search to determine the best configuration. The learning rate is chosen from $\{0.01, 0.02,0.025,0.05\}$. The training epoch is chosen from $\{2,3,4,5\}$. The batch size is chosen from $\{16,32,64\}$. The optimal parameters are task-specific.
For the other models, we adopt the best configurations reported in the source publications.

We report on statistical significance with a paired two-tailed t-test and we mark a significant improvement of \mtv over GPT-GNN for $p < 0.05$ with $^\blacktriangle$.

%% file: sections/05-results-and-analysis.tex

\section{Results and Analysis}
\label{sec:results-and-analysis}

We present the results of fine-tuning \mtv on four downstream tasks:
\begin{enumerate*}
\item link prediction,
\item similarity search,
\item node classification, and
\item node clustering.
\end{enumerate*}
We analyze the computational costs, conduct an ablation analysis, and study the parameter sensitivity.

\subsection{Downstream tasks}
\label{Sec:downstream task}
We evaluate PF-HIN over four downstream tasks, i.e., link prediction, similarity search, node classification  and node clustering.
\subsubsection{Link prediction}

This task is to predict which links would occur in the future. Unlike previous work~\cite{DBLP:conf/kdd/GroverL16} that randomly samples a certain percentage of links as the training dataset and uses the remaining links as the evaluation dataset, we adopt a sequential split of training and test data. We first train a binary logistic classifier on the graph of training data, and then use the test dataset with the same number of random negative (non-existent) links to evaluate the trained classifier. We only consider the new links in the training dataset and remove duplicate links from the evaluation. We adopt AUC and F1 score as evaluation metrics.

\begin{table} [t]
	\centering
	\caption{Results on the link prediction task.}
	\label{tab:link predict}
    \resizebox{\columnwidth}{!}{
    \setlength{\tabcolsep}{1mm}
		\begin{tabular}{lcccccccc}
			\toprule
			&\multicolumn{2}{c}{OAG-mini}&\multicolumn{2}{c}{DBLP}&\multicolumn{2}{c}{YELP}&\multicolumn{2}{c}{YAGO}\\
			\cmidrule(r){2-3}\cmidrule(lr){4-5}\cmidrule(lr){6-7}\cmidrule(l){8-9}
			Model & AUC & F1 & AUC & F1 & AUC & F1 & AUC & F1 \\
			\midrule
			DeepWalk     &0.378&0.266& 0.583 &0.351&0.602&0.467&0.735&0.525 \\
			LINE         &0.382&0.257& 0.274 &0.357&0.605&0.463&0.739&0.531  \\
			node2vec     &0.392&0.273& 0.584 &0.355&0.609&0.471&0.742&0.534  \\
			metapath2vec &0.412&0.282& 0.604 &0.367&0.618&0.473&0.744&0.541  \\
			HINE         &0.423&0.288& 0.607 &0.369&0.621&0.482&0.763&0.548 \\
			HIN2Vec      &0.426&0.291& 0.611 &0.376&0.625&0.493&0.768&0.578 \\
			GCN          &0.437&0.294& 0.623 &0.392&0.638&0.516&0.779&0.583 \\
			GraphSage    &0.445&0.293& 0.627 &0.395&0.641&0.525&0.783&0.592  \\
			GAT          &0.451&0.294& 0.631 &0.392&0.644&0.537&0.781&0.596 \\
			GIN          &0.456&0.299& 0.636 &0.394&0.647&0.539&0.785&0.598 \\
			HetGNN       &0.467&0.317& 0.642 &0.402&0.663&0.544&0.793&0.602 \\
			HGT        &0.473&0.321& 0.648 &0.407&0.672&0.549&0.799&0.605 \\
			GPT-GNN   &0.513&0.371& 0.678 &0.423&0.679&0.558&0.811&0.617 \\
			GCC       &0.507&0.352& 0.669 &0.417&0.668&0.552&0.805&0.609 \\
			MPP       &0.514&0.375& 0.671 &0.425&0.682&0.559&0.812&0.619 \\
						
			\midrule
			\mtv & \textbf{0.519}\rlap{$^\blacktriangle$} &\textbf{0.383}\rlap{$^\blacktriangle$} &\textbf{0.692}\rlap{$^\blacktriangle$} & \textbf{0.442}\rlap{$^\blacktriangle$} & \textbf{0.691}\rlap{$^\blacktriangle$} &\textbf{0.565}\rlap{$^\blacktriangle$} &\textbf{0.822}\rlap{$^\blacktriangle$} &\textbf{0.624}\rlap{$^\blacktriangle$} \\
			\bottomrule
		\end{tabular}%
}		
\end{table}

We present the link prediction results in Table~\ref{tab:link predict}, with the highest results set in bold.  Scores increase as the dataset size decreases. Traditional homogeneous models (DeepWalk, LINE, node2vec) perform worse than traditional heterogeneous metapath based models (metapath2vec, HINE, HIN2Vec); metapaths capture the network structure better than random walks. However, homogeneous GNN models (GCN, GraphSAGE, GAT, GIN) achieve even better results than traditional heterogeneous methods. Deep neural networks explore the entire network more effectively, generating better representations for link prediction. HetGNN and HGT outperform the homogeneous \ac{GNN} models, since they take the node types into consideration. GPT-GNN, GCC and MPP outperform all of the above methods including their base GNN (HGT and GIN). Adopting pre-training tasks can boost the downstream task performance.

\mtv outperforms GCC. This is because our pre-training on (type-based) mini-sequences helps to explore the HIN, while GIN is designed for homogeneous information. \mtv outperforms GPT-GNN due to our choice of pre-training task, as the ANP task is more effective on predicting links between nodes than the edge generation task used in GPT-GNN. By deciding whether two nodes are adjacent, ANP can tell if a link connects them. \mtv also outperforms MPP. This is because our heterogeneous node sequence contains more information than the meta-paths used by MPP.

\subsubsection{Similarity search}
In this task, we aim to find nodes that are similar to a given node. To evaluate the similarity between two nodes, we calculate the cosine similarity based on the node representations.  It is hard to rank all pairs of nodes explicitly, so we provide an estimation based on the grouping label $g(\cdot)$, in which similar nodes are gathered in one group.  Given a node $u$, if we rank other nodes based on the similarity score, intuitively, nodes from the same group (similar ones) should be at the top of the ranked list while dissimilar ones should be ranked at the bottom. We define the AUC value as:
\begin{equation}
\begin{split}
&\mathrm{AUC}={}\\
&\frac{1}{|V|}\sum_{u \in V} \frac{\sum\limits_{\substack{v,v' \in V \\ g(u)=g(v), g(u) \neq g(v') }} {^{\mathds{1}}\mathrm{sim}(u,v)>\mathrm{{sim}(u,v')}}}{\sum\limits_{\substack{v,v' \in V \\ g(u)=g(v), g(u) \neq g(v') }}1}.
\end{split}
\label{overall equation6}
\end{equation}

\begin{table} [t]
	\centering
	\caption{Results on the similarity search  task.}
	\label{tab:similarity search}	
		\begin{tabular}{lcccccc}
			\toprule
			&OAG-mini&DBLP&YELP&YAGO\\
			\cmidrule(r){2-5}
			Model & AUC  & AUC  & AUC  & AUC  \\
			\midrule
			DeepWalk     &0.478& 0.511 &0.553&0.656\\
			LINE         &0.482& 0.506 &0.558&0.661 \\
			node2vec     &0.483& 0.513 &0.559&0.653 \\
			metapath2vec &0.494& 0.545 &0.578&0.673 \\
			HINE         &0.506& 0.554 &0.583&0.683 \\
			HIN2Vec      &0.512& 0.556 &0.587&0.687 \\
			GCN          &0.509& 0.553 &0.581&0.682 \\
			GraphSage    &0.513& 0.557 &0.586&0.689 \\
			GAT          &0.510& 0.555 &0.584&0.691 \\
			GIN          &0.514& 0.559 &0.587&0.690 \\
			HetGNN       &0.527& 0.563 &0.592&0.694 \\
			HGT          &0.532& 0.568 &0.591&0.697 \\
			GPT-GNN      &0.563& 0.596 &0.603&0.708 \\
			GCC          &0.554& 0.587 &0.597&0.702 \\
			MPP          &0.559& 0.591 &0.593&0.705 \\
			
			\midrule
			\mtv & \textbf{0.574}\rlap{$^\blacktriangle$} &\textbf{0.612}\rlap{$^\blacktriangle$} &\textbf{0.613}\rlap{$^\blacktriangle$} & \textbf{0.719}\rlap{$^\blacktriangle$}  \\
			\bottomrule
		\end{tabular}%
			
\end{table}

\noindent%
Table~\ref{tab:similarity search} displays the results on the similarity search task. The highest scores are set in bold. The traditional heterogeneous models and homogeneous GNN models achieve comparable results; metapath based mechanisms and deep neural networks can generate expressive node embeddings for similarity search. HetGNN and HGT outperform the above methods, which shows the power of combining GNN and type features. GPT-GNN is the best performing baseline as it leverages both pre-training power and type features.  \mtv performs the best in all cases, illustrating the effectiveness of our pre-training and fine-tuning framework on learning heterogeneous node representations for similarity search.

\subsubsection{Node classification}
Next we report on the results for the multi-label node classification task. The size (ratio) of the training data is set to 30\% and the remaining nodes are used for testing. 
We adopt micro-F1 (MIC) and macro-F1 (MAC) as our evaluation metrics.

\begin{table} [htbp]
	\centering
	\caption{Results on the multi-label node classification task.}
	\label{tab:node classification}
	\resizebox{\columnwidth}{!}{
	\setlength{\tabcolsep}{3pt}
		\begin{tabular}{@{}l cccccccc}
			\toprule
			&\multicolumn{2}{c}{OAG-mini}&\multicolumn{2}{c}{DBLP}&\multicolumn{2}{c}{YELP}&\multicolumn{2}{c}{YAGO}\\
			\cmidrule(r){2-3}\cmidrule(lr){4-5}\cmidrule(lr){6-7}\cmidrule(l){8-9}
			Model & MIC &  MAC & MIC& MAC & MIC & MAC& MIC&  MAC\\
			\midrule
			DeepWalk     &0.175 &0.173& 0.193 &0.191&0.163&0.145&0.328&0.265\\
			LINE         &0.177 &0.172&0.184 &0.179&0.274&0.276&0.366&0.320 \\
			node2vec     &0.180 &0.178&0.201 &0.198&0.194&0.151&0.332&0.280 \\
			metapath2vec &0.195 &0.194&0.209 &0.207&0.264&0.269&0.370&0.332 \\
			HINE         &0.198 &0.192&0.234 &0.230&0.276&0.284&0.401&0.363 \\
			HIN2Vec      &0.204 &0.201&0.246 &0.241&0.291&0.306&0.428&0.394 \\
			GCN          &0.209 &0.205&0.257 &0.256&0.302&0.311&0.459&0.447 \\
			GraphSage    &0.211 &0.207&0.267 &0.269&0.305&0.318&0.464&0.456 \\
			GAT          &0.213 &0.211&0.271 &0.273&0.303&0.315&0.469&0.462\\
			GIN          &0.218 &0.223&0.274 &0.277&0.308&0.321&0.474&0.466 \\
			HetGNN       &0.234 &0.237&0.285 &0.282&0.309&0.324&0.478&0.471\\
			HGT          &0.239 &0.241&0.283 &0.278&0.313&0.327&0.484&0.483\\
			GPT-GNN      &0.276 &0.274&0.304 &0.299&0.322&0.339&0.496&0.493\\
			GCC          &0.266 &0.264&0.295 &0.291&0.315&0.332&0.489&0.486\\
			MPP          &0.269 &0.268&0.298 &0.294&0.318&0.336&0.493&0.488\\

			\midrule
			\mtv & \textbf{0.294}\rlap{$^\blacktriangle$} &\textbf{0.292}\rlap{$^\blacktriangle$} &\textbf{0.318}\rlap{$^\blacktriangle$} & \textbf{0.311}\rlap{$^\blacktriangle$} & \textbf{0.330}\rlap{$^\blacktriangle$} &\textbf{0.354}\rlap{$^\blacktriangle$} &\textbf{0.516}\rlap{$^\blacktriangle$} &\textbf{0.509}\rlap{$^\blacktriangle$} \\
			\bottomrule
		\end{tabular}%
	}		
\end{table}

Table~\ref{tab:node classification} provides the results on the node classification task; the highest scores are set in bold. GNN based models perform relatively well, showing the benefits of using deep neural networks for exploring features of the network data for classification. \mtv achieves the highest scores thanks to our fine-tuning framework, which aggregates the full sequence information for node classification.

\subsubsection{Node clustering}
\label{sec:node cluster}

Finally, we report on the outcomes on the node clustering task. We feed the generated node embeddings of each model into a clustering model. Here, we choose a k-means algorithm to cluster the data. The size (ratio) of the training data is set to 30\% and the remaining nodes are used for testing. We use  normalized mutual information (NMI) and adjusted rand index (ARI) as evaluation metrics.

\begin{table} [htbp]
	\centering
	\caption{Results on the node clustering task.}
	\label{tab:node cluster}
	\resizebox{\columnwidth}{!}{%
	\setlength{\tabcolsep}{3pt}
		\begin{tabular}{l@{~}cccccccc}
			\toprule
			&\multicolumn{2}{c}{OAG-mini}&\multicolumn{2}{c}{DBLP}&\multicolumn{2}{c}{YELP}&\multicolumn{2}{c}{YAGO}\\
			\cmidrule(r){2-3}\cmidrule(lr){4-5}\cmidrule(lr){6-7}\cmidrule(l){8-9}
			Model & NMI &  ARI & NMI &  ARI & NMI &  ARI& NMI &  ARI\\
			\midrule
			DeepWalk      &0.601 &0.613& 0.672 &0.686&0.713& 0.744 & 0.856 &0.886  \\
			LINE          &0.598 &0.609&0.678 &0.693&0.705& 0.739 & 0.861 &0.894 \\
			node2vec      &0.604 &0.616&0.673 &0.689&0.719& 0.748 & 0.867 &0.899 \\
			metapath2vec &0.628 &0.632&0.711 &0.738&0.748&0.785&0.896&0.917\\
			HINE         &0.634 &0.638&0.718 &0.741&0.753&0.786&0.899&0.921\\
			HIN2Vec      &0.637 &0.640&0.721 &0.744&0.749&0.789&0.902&0.923\\
			GCN          &0.616 &0.641&0.701 &0.719&0.744&0.775&0.881&0.907\\
			GraphSage    &0.619 &0.643&0.705 &0.722&0.746&0.778&0.885&0.911\\
			GAT          &0.622 &0.645&0.709 &0.728&0.748&0.782&0.893&0.915\\
			GIN          &0.625 &0.650&0.711 &0.735&0.752&0.785&0.898&0.919\\
			HetGNN       &0.644 &0.678&0.729 &0.748&0.759&0.792&0.904&0.926\\
			HGT          &0.648 &0.682&0.734 &0.753&0.763&0.794&0.909&0.931\\
			GPT-GNN      &0.673 &0.712&0.756 &0.761&0.769&0.808&0.919&0.937\\
			GCC          &0.652 &0.707&0.741 &0.756&0.765&0.799&0.913&0.933\\
			MPP          &0.662 &0.709&0.747 &0.758&0.766&0.802&0.915&0.935\\
				
			\midrule
			\mtv & \textbf{0.691}\rlap{$^\blacktriangle$}&\textbf{0.728}\rlap{$^\blacktriangle$}&\textbf{0.770}& \textbf{0.773}\rlap{$^\blacktriangle$} & \textbf{0.781}\rlap{$^\blacktriangle$}&\textbf{0.816}\rlap{$^\blacktriangle$} &\textbf{0.927}\rlap{$^\blacktriangle$} &\textbf{0.946}\rlap{$^\blacktriangle$} \\
			\bottomrule
		\end{tabular}
	}
\end{table}

Table~\ref{tab:node cluster} shows the performance on the node clustering task, with the highest scores set in bold. Despite the strong ability of homogeneous GNN models to capture structural information of a network, they perform slightly worse than traditional heterogeneous models. Taking node type information into account makes a real difference when clustering nodes. GPT-GNN is the best performing baseline, combining type information and pre-training power.  \mtv outperforms all the baselines, proving that \mtv is able to generate effective node embeddings for node clustering.

\subsection{Computational costs}
	To evaluate the efficiency of our fine-tuning framework compared to other models, we conduct an analysis of the computational costs. Specifically, we analyze the running time of each model on each task, using the early stopping mechanism.
Due to space limitations we only report the results on the DBLP dataset; the results for the remaining datasets are qualitatively similar.
See Table~\ref{tab:computation cost}.
We use standard hardware (Intel (R) Core (TM) i7-10700K CPU +  GTX-2080 GPU); the time reported is wall-clock time, averaged over 10 runs.

\begin{table} [t]
	\centering
	\caption{Running times 
on the DBLP dataset. Abbreviations used: LP: link prediction, SS: similarity search, MC: multi-label node classification, and NC: node clustering.}
	\label{tab:computation cost}
	\begin{tabular}{l rrrr}
		\toprule
		&LP&SS&MC&NC\\
		\cmidrule(r){2-5}
		Model & Time (h)  & Time (h)  &  Time (h)   &  Time (h)   \\
		\midrule
		DeepWalk     & 1.09 &1.15&0.74&0.53\\
		LINE         & 1.19 &1.31&0.76&0.62 \\
		node2vec     & 1.46 &1.54&0.97&0.74 \\
		metapath2vec & 1.68 &1.90&1.35&1.05 \\
		HINE         & 1.84 &2.11&1.38&1.08 \\
		HIN2Vec      & 2.02 &2.53&1.56&1.26 \\
		GCN          & 3.45 &4.14&2.66&2.08 \\
		GraphSage    & 2.75 &3.25&2.17&1.46 \\
		GAT          & 2.55 &2.94&1.73&1.55 \\
		GIN          & 3.59 &4.54&3.21&2.33 \\
		HGT          & 3.11 &3.80&2.53&2.19\\
		HetGNN       & 3.15 &3.68&2.51&1.80 \\
		GPT-GNN      & 1.64 &1.92&1.35&1.03 \\
		GCC          & 1.55 &2.28&1.56&1.13 \\	
		\midrule
		\mtv & 1.52 & 1.71 &1.18 & 0.82  \\
		\bottomrule
	\end{tabular}%
	
\end{table}

\mtv's running time is longer than of the three traditional homogeneous models DeepWalk, LINE and node2vec, which are based on  random walks; it is not as high as any of the other models. GNN based models like GCN, GraphSAGE, GAT, GIN and HetGNN have a higher running time than all other models, since the complexity of traditional deep neural networks is much higher than other algorithms. GPT-GNN, GCC and \mtv have relatively short running times; this is because pre-trained parameters help the loss function converge much faster. Among the pre-train and fine-tune models, \mtv is the most efficient one since the complexity of the transformer encoder we use is lower than that of HGT in GPT-GNN and that of GIN in GCC.

\subsection{Ablation analysis} \label{sec:ablation}
We analyze the effect of the pre-training tasks, the bi-directional transformer encoders, the components of the input representation, and the rank-guided heterogeneous walk sampling strategy.

\subsubsection{Effect of pre-training task}
To evaluate the effect of the pre-training tasks, we introduce two variants of \mtv, i.e., \mtv{}$\setminus$MNM and \mtv{}$\setminus$ANP. \mtv{}$\setminus$MNM is like \mtv but excludes pre-training on the masked node modeling task,  \mtv{}$\setminus$ANP is like \mtv but excludes pre-training on the adjacent node prediction task. Due to space limitations, we only report the experimental outcomes on the DBLP dataset; the findings on the other datasets are qualitatively similar.
Table~\ref{tab:pre-training} shows the experimental results of the ablation analysis over pre-training tasks. For the link prediction task, ANP is more important than MNM since predicting if two nodes are adjacent could also tell if they are connected by a link. In similarity search, those two tasks have a comparable effect. For the node classification and node clustering tasks, MNM plays a more important role, and this is because MNM directly models node features and hence \mtv is better able to explore them.

\begin{table} [t]
	\centering
	\caption{Ablation analysis of the pre-training tasks on the DBLP dataset. Same abbreviations used as in Table~\ref{tab:computation cost}.}
	\label{tab:pre-training}
	\resizebox{\columnwidth}{!}{%
	\setlength{\tabcolsep}{3pt}
		\begin{tabular}{@{} l@{~}ccccccc @{}}
			\toprule
			&\multicolumn{2}{c}{LP}&SS&\multicolumn{2}{c}{MC}&\multicolumn{2}{c}{NC}\\
			\cmidrule(r){2-3}\cmidrule(lr){4-4}\cmidrule(lr){5-6}\cmidrule(l){7-8}
			Model & AUC &  F1 &  AUC & MIC &  MAC& NMI &  ARI\\
			\midrule
			\mtv      & \textbf{0.692} & \textbf{0.442} & \textbf{0.612} & \textbf{0.318} & \textbf{0.311} & \textbf{0.770} & \textbf{0.773} \\
			\mtv{}$\setminus$ANP  & 0.421 &0.221&0.356&0.217&0.257&0.548&0.629 \\
			\mtv{}$\setminus$MNM  & 0.498 &0.283&0.332&0.172&0.174&0.422&0.438 \\
			\bottomrule
		\end{tabular}%
	}
\end{table}

\subsubsection{Effect of the bi-directional transformer encoder}
Our bi-direc\-tional transformer encoder is a variant of GNN applied to \acp{HIN}, aggregating  neighborhood information. For our analysis, we replace the transformer encoders with a CNN,  a bi-directional LSTM, and an attention mechanism. Specifically, the model using the CNN encoder is denoted as \mtv{}(CNN), the model using bi-directional LSTM as \mtv{}(LSTM), and the model using an attention mechanism as \mtv{}(attention). Again, we report on experiments on the DBLP dataset only.
Table~\ref{tab:transfomer} presents the experimental results of different encoders. The CNN, LSTM and attention mechanism based models achieve comparable results on the four tasks. \mtv consistently  outperforms all three models, which shows the importance of our bi-directional transformer encoder for mining the information contained in a \ac{HIN}.

\begin{table} [t]
	\centering
	\caption{Ablation analysis of the encoder on the DBLP dataset. Same abbreviations used as in Table~\ref{tab:computation cost}.}
	\label{tab:transfomer}
	\resizebox{\columnwidth}{!}{%
	\setlength{\tabcolsep}{3pt}
		\begin{tabular}{@{}l@{~}ccccccc@{}}
			\toprule
			&\multicolumn{2}{c}{LP}&SS&\multicolumn{2}{c}{MC}&\multicolumn{2}{c}{NC}\\
			\cmidrule(r){2-3}\cmidrule(lr){4-4}\cmidrule(lr){5-6}\cmidrule(l){7-8}
			Model & AUC &  F1 &  AUC & MIC &  MAC& NMI &  ARI\\
			\midrule
			\mtv      & \textbf{0.692} & \textbf{0.442} & \textbf{0.612} & \textbf{0.318} & \textbf{0.311} & \textbf{0.770} & \textbf{0.773} \\
			\mtv{}(CNN)       &0.643&0.418&0.578&0.296&0.287&0.743&0.747 \\
			\mtv{}(LSTM)     &0.657&0.421&0.567&0.293&0.279&0.754&0.753 \\
			\mtv{}(attention) &0.629&0.411&0.572&0.299&0.291&0.748&0.757 \\
			\bottomrule
		\end{tabular}%
	}
\end{table}

\subsubsection{Effect of pre-training strategy}
In \mtv, the MNM task is pre-trained on (type-based) mini-sequences, while the ANP task is pre-trained on two full sequences generated via two starting nodes. Here we analyze the effect of this strategy. We consider three variants of \mtv. The first is to pre-train the two tasks on two full sequences without considering the heterogeneous features of the network, denoted as \mtv{}(full-full). In the second, we try to assign the ANP task with a sense of heterogeneous features. As explained in Section~\ref{ANP}, it is too time-consuming to pre-train all mini-sequence pairs for the ANP task, so in the second model we choose the two longest mini-sequences to train the ANP task, while the MNM task is trained on full sequences, denoted as \mtv{}(full-mini). The third is that the ANP and MNM tasks are both trained on (type-based) mini-sequences, denoted as  \mtv{}(mini-mini).
Table~\ref{tab:pretrain} shows the results of the ablation analysis. \mtv{}(full-full) outperforms \mtv{}(full-mini), which illustrates that despite taking heterogeneous features into consideration, only choosing two mini-sequences for the ANP task may harm the performance as information is missed. However, \mtv{}(mini-mini) outperforms \mtv{}(full-full), showing that considering heterogeneous features on the MNM task may help boost the model performance. The strategy selected for \mtv achieves the best results.

\begin{table} [t]
 	\small
 	\centering
 	\caption{Ablation analysis of the pre-training strategy on the DBLP dataset. Same abbreviations used as in Table~\ref{tab:computation cost}.}
 	\label{tab:pretrain}
 	\resizebox{\columnwidth}{!}{%
	\setlength{\tabcolsep}{3pt}
 		\begin{tabular}{l@{~}ccccccc}
 			\toprule
 			&\multicolumn{2}{c}{LP}&SS&\multicolumn{2}{c}{MC}&\multicolumn{2}{c}{NC}\\\cmidrule(r){2-3}\cmidrule(lr){4-4}\cmidrule(lr){5-6}\cmidrule(l){7-8}
 			Model & AUC &  F1 &  AUC & MIC &  MAC& NMI &  ARI\\
 			\midrule
 			\mtv      & \textbf{0.692} & \textbf{0.442} & \textbf{0.612} & \textbf{0.318} & \textbf{0.311} & \textbf{0.770} & \textbf{0.773} \\
 			\mtv{}(full-full) &0.671&0.419&0.589&0.295&0.294&0.747&0.757 \\
 			\mtv{}(full-mini) &0.659&0.407&0.577&0.287&0.283&0.733&0.751 \\
 			\mtv{}(mini-mini) &0.679&0.429&0.596&0.307&0.301&0.759&0.766 \\
 			\bottomrule
 		\end{tabular}%
 	}
\end{table}

\begin{table} [htbp]
	\centering
	\caption{Ablation analysis of the rank-guided heterogeneous walk sampling strategy on the DBLP dataset. Same abbreviations used as in Table~\ref{tab:computation cost}.}
	\label{tab:ranking-based}
	\resizebox{\columnwidth}{!}{%
		\setlength{\tabcolsep}{3pt}
		\begin{tabular}{l@{~}ccccccc}
			\toprule
			&\multicolumn{2}{c}{LP}&SS&\multicolumn{2}{c}{MC}&\multicolumn{2}{c}{NC}\\
			\cmidrule(r){2-3}\cmidrule(lr){4-4}\cmidrule(lr){5-6}\cmidrule(l){7-8}
			Model & AUC &  F1 &  AUC & MIC & MAC& NMI & ARI\\
			\midrule
			\mtv      & \textbf{0.692} & \textbf{0.442} & \textbf{0.612} & \textbf{0.318} & \textbf{0.311} & \textbf{0.770} & \textbf{0.773}  \\
			\mtv{}(BFS)       &0.659&0.417&0.577&0.279&0.292&0.743&0.758 \\
			\mtv{}(DFS)       &0.643&0.406&0.565&0.269&0.283&0.741&0.752 \\
			\mtv{}(random)    &0.668&0.423&0.589&0.292&0.297&0.657&0.764 \\
			\bottomrule
		\end{tabular}%
	}
\end{table}

\subsubsection{Effect of rank-guided heterogeneous walk sampling}
In this paper, we have adopted a rank-guided heterogeneous walk sampling strategy to sample nodes to form input sequences. Here we introduce three variants. The first is to only use a breadth first search (BFS) sampling strategy, denoted as \mtv{}(BFS); the second is to only use a depth first search (DFS) sampling strategy, denoted as \mtv{}(DFS); and the last is to randomly choose neighboring nodes to form the node sequence, denoted as \mtv{}(random).

Table~\ref{tab:ranking-based} shows the experimental results. \mtv{}(BFS) outperforms \mtv{}(DFS) and \mtv{}(random); aggregating a node's closest neighborhood's information is more informative than choosing far-away nodes or randomly chosen neighboring nodes. \mtv outperforms \mtv{}(BFS); choosing nodes with a higher importance leads to better performing feature representations of a HIN.
\subsubsection{Effect of centrality metric}
In this paper, we have three centrality metrics to measure the importance of a node. Here we introduce three variants. The first is to remove the betweenness centrality denoted as \mtv{}$\setminus$betweenness; the second is to remove the eigencentrality denoted as \mtv{}$\setminus$eigen; the third is to remove the closeness centrality denoted as \mtv{}$\setminus$closeness. Each variant assigns equal weights for the left two metrics.

\begin{table} [ht]
	\centering
	\caption{Ablation analysis of the centrality metric on the DBLP dataset. Same abbreviations used as in Table~\ref{tab:computation cost}.}
	\label{tab:centrality}
	\resizebox{\columnwidth}{!}{%
		\setlength{\tabcolsep}{3pt}
		\begin{tabular}{l@{~}ccccccc}
			\toprule
			&\multicolumn{2}{c}{LP}&SS&\multicolumn{2}{c}{MC}&\multicolumn{2}{c}{NC}\\
			\cmidrule(r){2-3}\cmidrule(lr){4-4}\cmidrule(lr){5-6}\cmidrule(l){7-8}
			Model & AUC &  F1 &  AUC & MIC & MAC& NMI & ARI\\
			\midrule
			\mtv      & \textbf{0.692} & \textbf{0.442} & \textbf{0.612} & \textbf{0.318} & \textbf{0.311} & \textbf{0.770} & \textbf{0.773} \\
			\mtv{}$\setminus$betweenness       &0.674&0.423&0.582&0.304&0.298&0.757&0.759 \\
			\mtv{}$\setminus$eigen       &0.668&0.416&0.586&0.297&0.292&0.753&0.751 \\
			\mtv{}$\setminus$closeness   &0.671&0.420&0.577&0.301&0.302&0.761&0.754 \\
			\bottomrule
		\end{tabular}%
	}
\end{table}

Table~\ref{tab:centrality} presents the experimental results. Removing a metric will influence the experimental results which illustrates that each metric is necessary for ranking the nodes. In addition, each metric has different influence on different tasks, so it is reasonable for us to adopt the learnable weights on them.

\subsection{Parameter sensitivity}

\begin{figure}[h]
	\centering
	\subfigure[Link prediction]{
		\label{fig:length-lr}
		\includegraphics[clip,trim=1mm 2mm 9mm 2mm,width=0.35\textwidth]{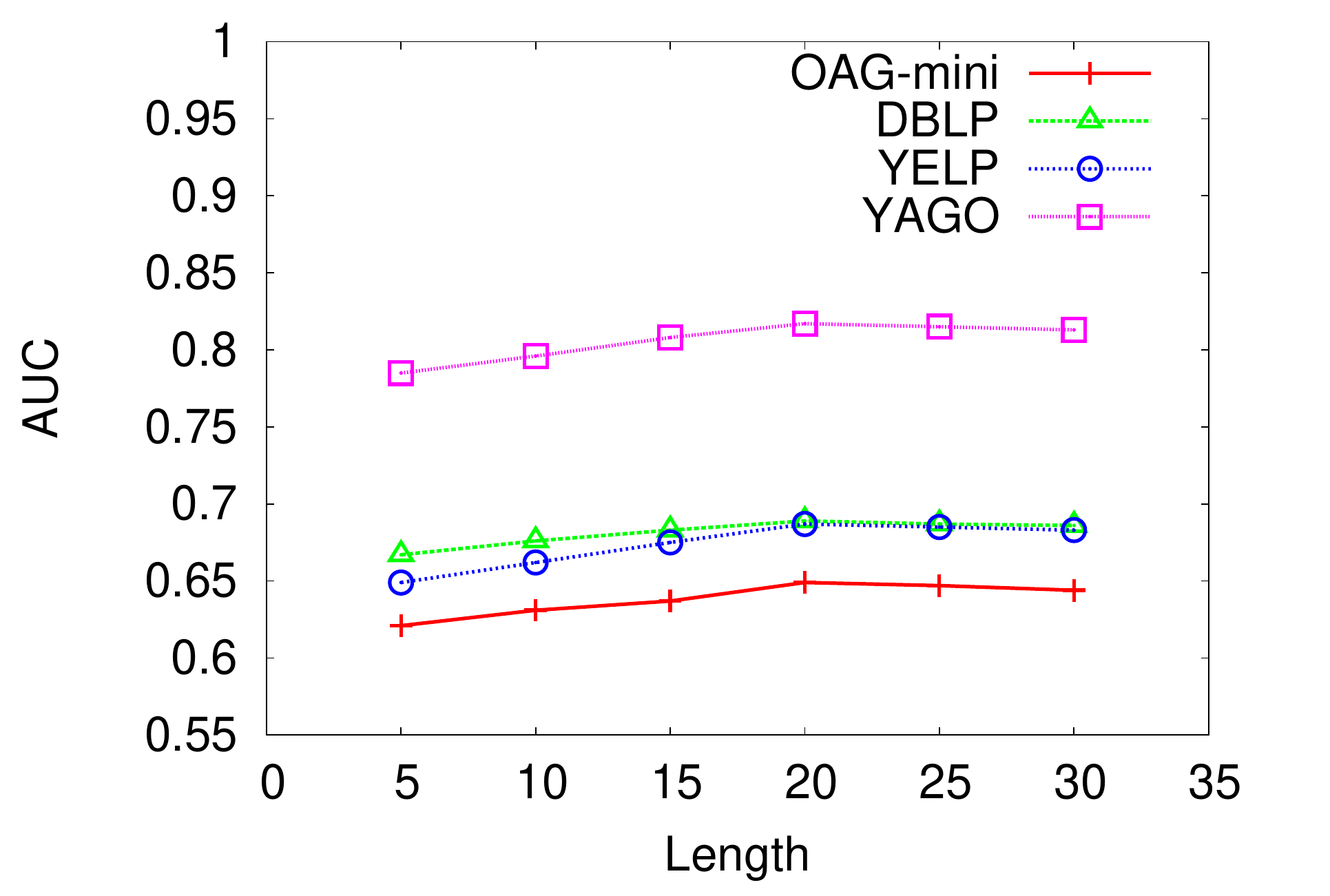}
	}
	\subfigure[Similarity search]{
		\label{fig:length-ss}
		\includegraphics[clip,trim=1mm 2mm 9mm 2mm,width=0.35\textwidth]{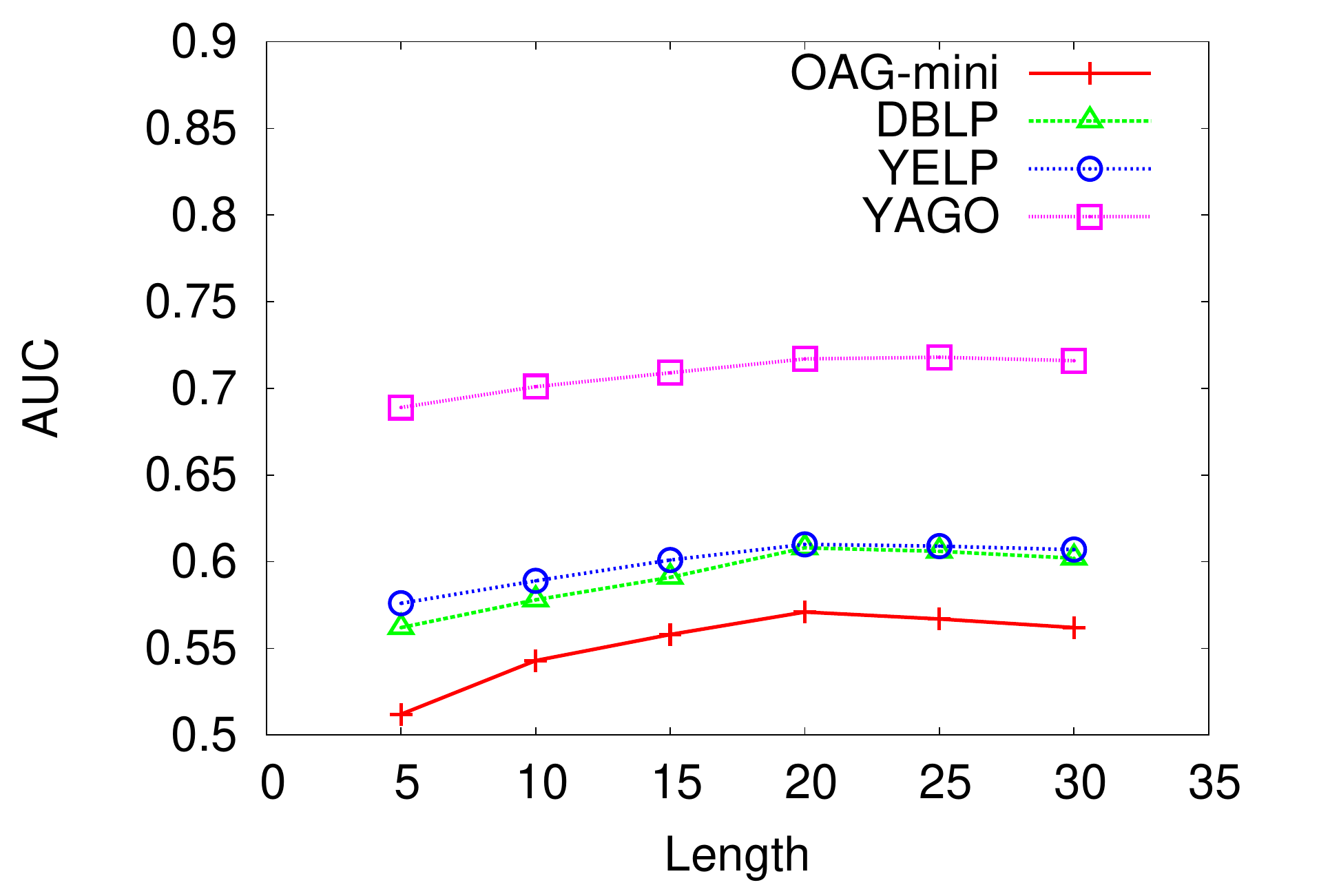}
	}	
	\\[-1ex]
	\subfigure[Node classification]{
		\label{fig:length-mc}
		\includegraphics[clip,trim=1mm 2mm 9mm 2mm,width=0.35\textwidth]{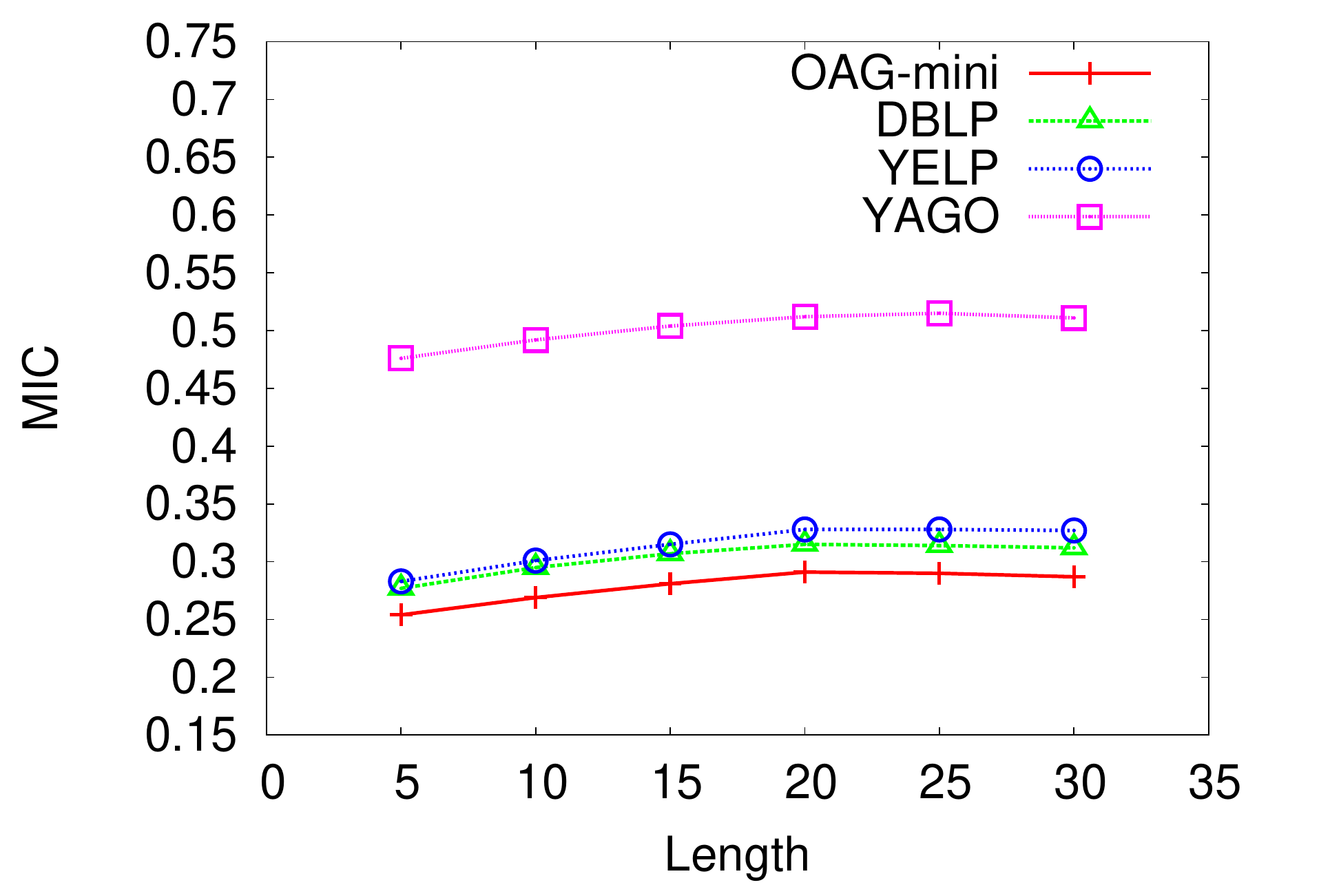}
	}
	\subfigure[Node clustering]{
		\label{fig:length-nc}
		\includegraphics[clip,trim=1mm 2mm 9mm 2mm,width=0.35\textwidth]{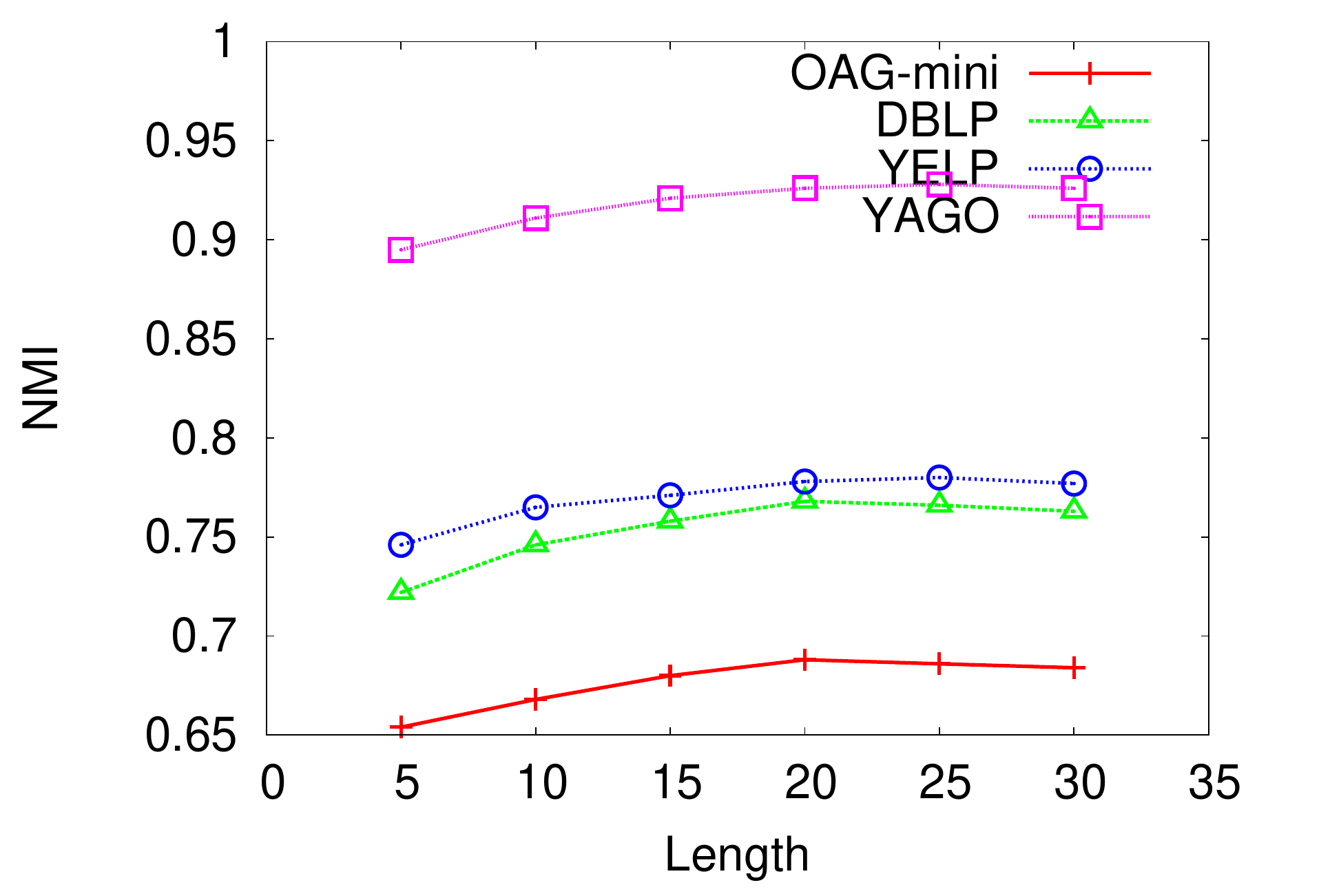}
	}
	
	\caption{Sensitivity w.r.t.\ maximum length of sequences.}
	\label{fig:length}
	
\end{figure}

\begin{figure}[h]
	\centering
	\subfigure[Link prediction]{
		\label{fig:dimension-lr}
		\vspace*{-1mm}\includegraphics[clip,trim=1mm 2mm 9mm 2mm,width=0.35\textwidth]{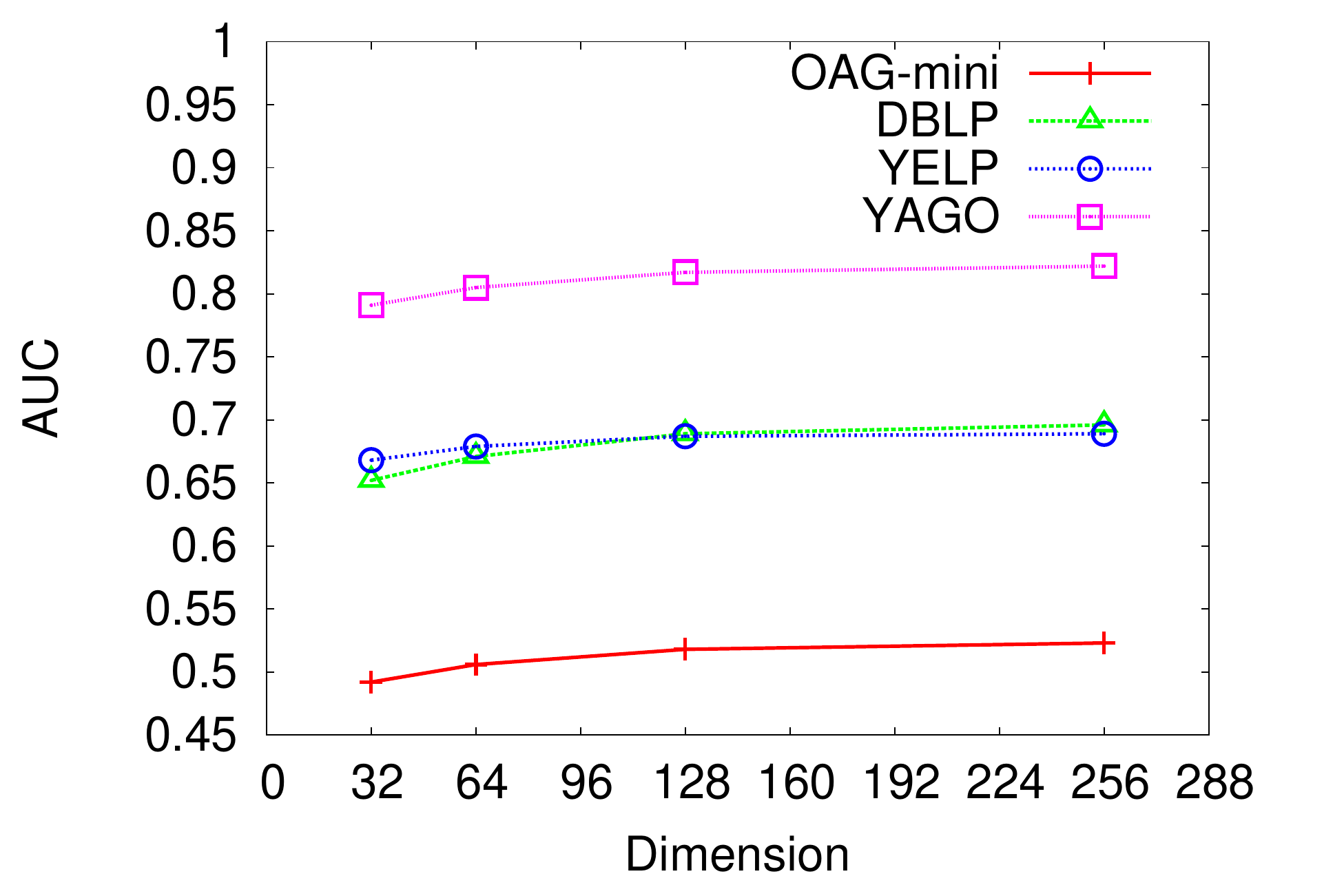}\vspace*{-1mm}
	}
	\subfigure[Similarity search]{
		\label{fig:dimension-ss}
		\includegraphics[clip,trim=1mm 2mm 9mm 2mm,width=0.35\textwidth]{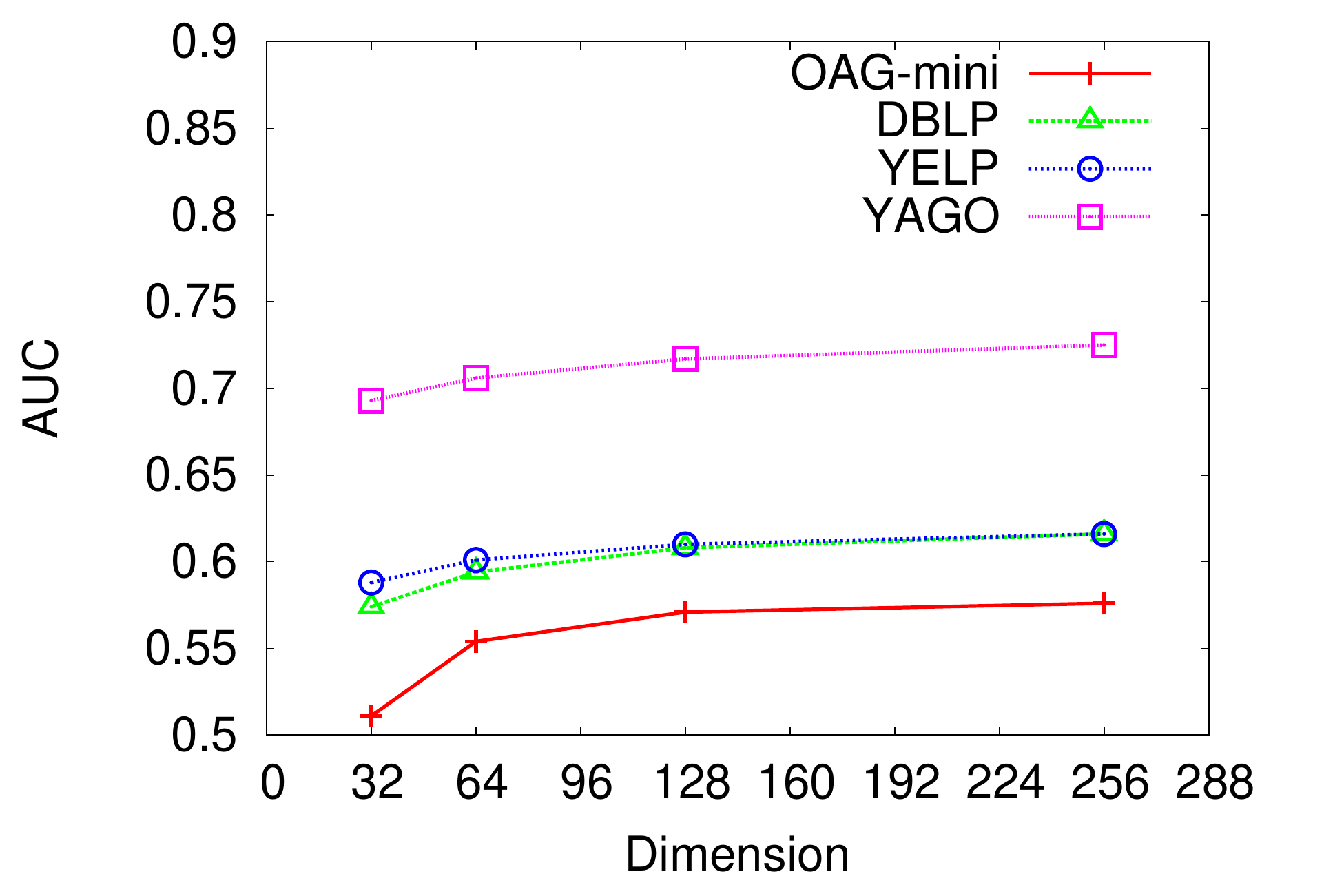}
	}
	\\[-1ex]
	\subfigure[Node classification]{
		\label{fig:dimension-mc}
		\vspace*{-10mm}\includegraphics[clip,trim=1mm 2mm 9mm 2mm,width=0.35\textwidth]{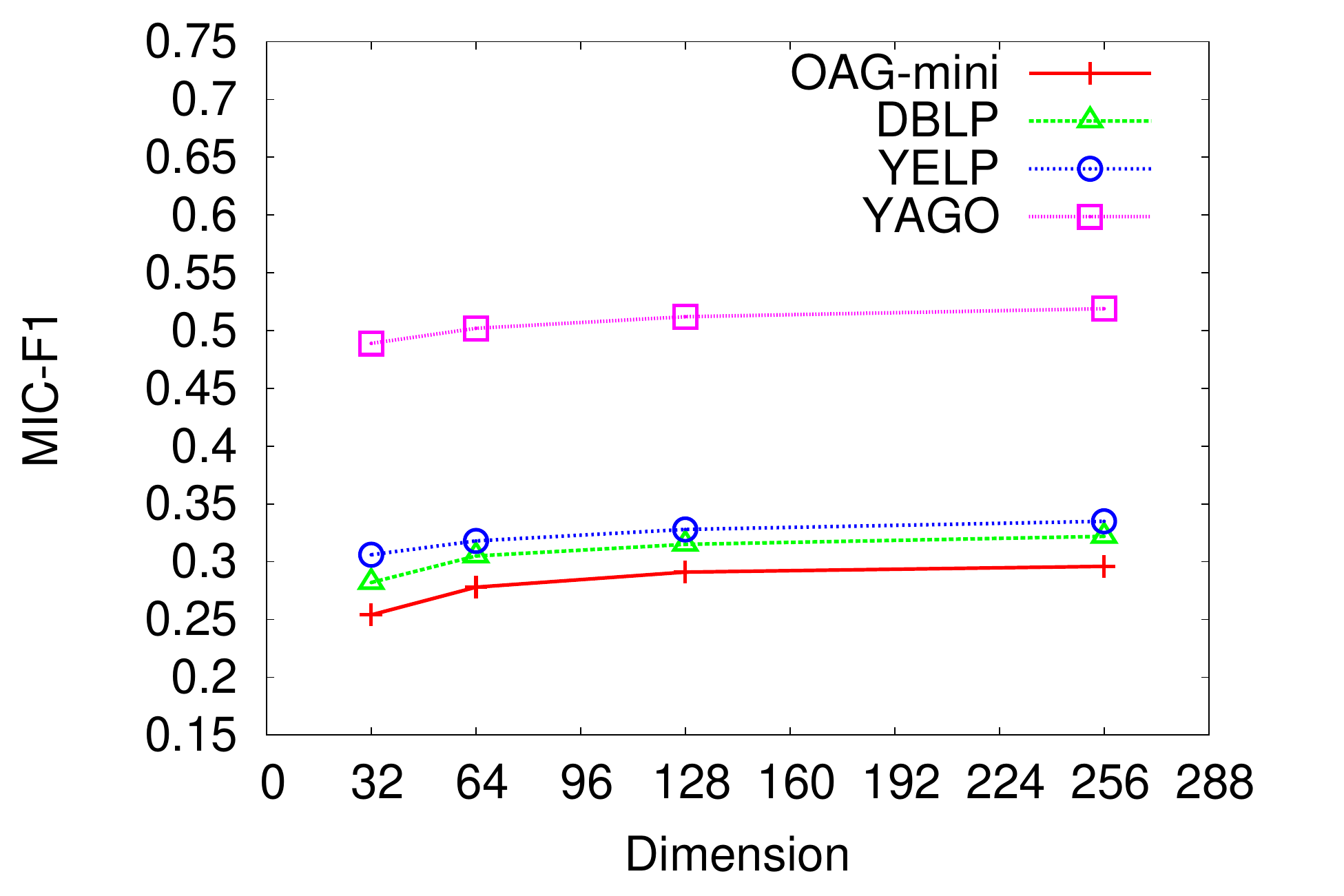}\vspace*{-1mm}
	}
	\subfigure[Node clustering]{
		\label{fig:dimension-nc}
		\includegraphics[clip,trim=1mm 2mm 9mm 2mm,width=0.35\textwidth]{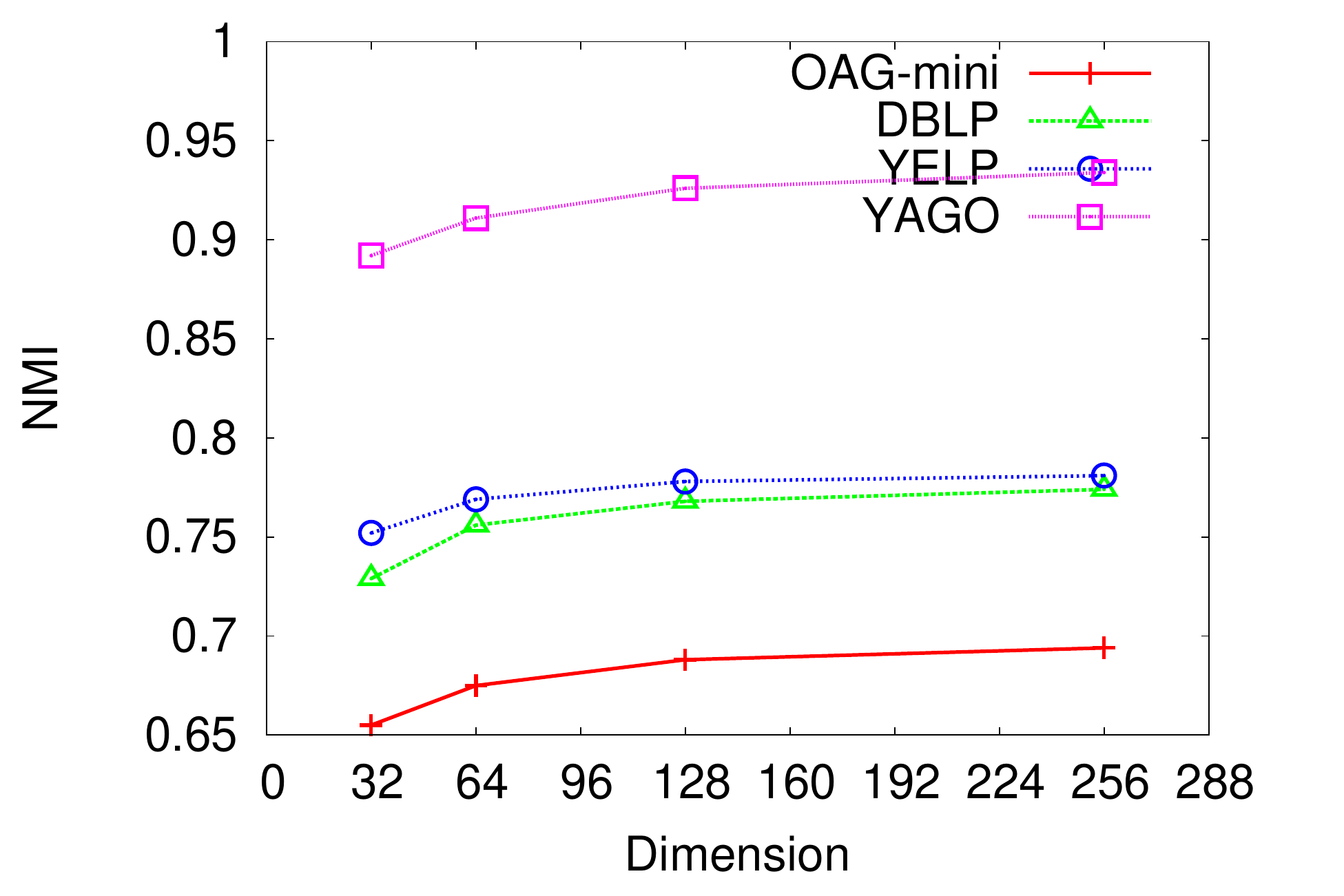}
	}
	
	\caption{Sensitivity w.r.t.\ dimension of node embeddings.}
	\label{fig:dimension}
	
\end{figure}

We conduct a sensitivity analysis of the hyper-parameters of \mtv. We choose two parameters for analysis: the maximum length of the input sequence and the dimension of the node embedding. For each downstream task, we only choose one metric for evaluation: AUC for link prediction, AUC for similarity search, micro-F1 value for node classification, and NMI value for node clustering.  Figures~\ref{fig:length} and~\ref{fig:dimension} show the results of our parameter analysis.

Figure~\ref{fig:length} shows the results for the maximum length of the input sequence. The performance improves rapidly when the length increases from 0 to 20. A short node sequence is not able to fully express neighborhood information. When the length reaches 20 or longer, the performance stabilizes, and longer sequence lengths may even hurt the performance. Given a node, its neighboring information can be well represented by its direct neighborhood, however, including far-away nodes may introduce noise. According to this analysis, we choose the length of the input sequence to be 20 so as to balance effectiveness and efficiency.

As to the dimension of node embeddings, Figure~\ref{fig:dimension} shows that the performance improves as the dimension increases, for all tasks and datasets. The higher dimensions are able to capture more features. \mtv is not very sensitive to the dimension we choose, especially once it is at least 128. The performance gap is not very large between dimension 128 and 256. Thus, we choose 128 as our setting for the dimension of node embeddings for efficiency considerations.

%% file: sections/06-conclusion.tex

\section{Conclusions}
\label{sec:conclusion}

We have considered the problem of network representation learning for \acfp{HIN}.
We propose a novel model, \mtv, to mine the information captured by a HIN. \mtv is a self-supervised pre-training and fine-tuning framework. In the pre-training stage, we first use rank-guided heterogeneous walks to generate input sequences and group them into (type-based) mini-sequences. The pre-training tasks we utilize are masked node modeling (MNM) and adjacent node prediction (ANP). Then we leverage bi-directional transformer layers to pre-train the model. We adopt factorized embedding parameterization and cross-layer parameter sharing strategies to reduce the number of parameters. We fine-tune \mtv on four tasks: link prediction, similarity search, node classification, and node clustering. \mtv significantly and consistently outperforms state-of-the-art models on the above tasks on four real-life datasets.

In future work, we plan to conduct further graph learning tasks in the context of a diverse range of information retrieval tasks, including but not limited to academic search, financial search, product search, and social media search. It is also of interest to see how to model a dynamic HIN that is constantly evolving, using a pre-training and fine-tuning framework.